\documentclass{article}

\usepackage{template/arxiv}
\usepackage[utf8]{inputenc}
\usepackage[T1]{fontenc}
\usepackage[colorlinks=false, hidelinks]{hyperref}
\usepackage{url}
\usepackage{latexsym}
\usepackage{amsfonts}
\usepackage{amsmath}
\usepackage{amsthm}
\usepackage{booktabs}
\usepackage{enumitem}
\usepackage{nicefrac}
\usepackage{microtype}
\usepackage{cleveref}
\usepackage{graphicx}
\usepackage{natbib}
\usepackage{subcaption}
\usepackage{float}
\usepackage{xcolor}
\usepackage{svg}
\usepackage{doi}
\usepackage{lipsum}

\title{Hybrid-AIRL: Enhancing Inverse Reinforcement Learning with Supervised Expert Guidance}

\usepackage{authblk}

\newbox{\orcid}\sbox{\orcid}{\includegraphics[scale=0.06]{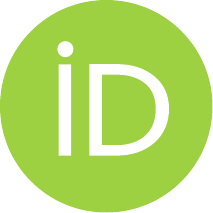}}

\author[1]{%
    \href{https://orcid.org/0009-0004-1465-7772}{\usebox{\orcid}\hspace{1mm}
    Bram Silue}%
}
\author[1]{%
    \hspace{1px}
    Santiago Amaya-Corredor
}
\author[2]{%
    \href{https://orcid.org/0000-0002-7951-878X}{\usebox{\orcid}\hspace{1mm}
    Patrick Mannion}%
}
\author[3]{%
    \href{https://orcid.org/0000-0002-9210-1196}{\usebox{\orcid}\hspace{1mm}
    Lander Willem}%
}
\author[1]{%
    \href{https://orcid.org/0000-0003-3906-758X}{\usebox{\orcid}\hspace{1mm}
    Pieter Libin}%
}

\affil[1]{Artificial Intelligence Lab — Vrije Universiteit Brussel}
\affil[2]{College of Science and Engineering — University of Galway}
\affil[3]{Family Medicine and Population Health — University of Antwerp}

\renewcommand{\shorttitle}{Hybrid Adversarial Inverse Reinforcement Learning}

\hypersetup{
    pdftitle={Hybrid-AIRL: Enhancing Inverse Reinforcement Learning with Supervised Expert Guidance},
    pdfsubject={cs.AI, cs.LG},
    pdfauthor={Bram Silue, Santiago Amaya-Corredor, Patrick Mannion, Lander Willem, Pieter Libin},
    pdfkeywords={Inverse Reinforcement Learning, Supervised Learning, Generative Adversarial Networks, Poker}
}

\begin{document}
\maketitle

\begin{abstract}
Adversarial Inverse Reinforcement Learning (AIRL) has shown promise in addressing the sparse reward problem in reinforcement learning (RL) by inferring dense reward functions from expert demonstrations. However, its performance in highly complex, imperfect-information settings remains largely unexplored. To explore this gap, we evaluate AIRL in the context of Heads-Up Limit Hold'em (HULHE) poker, a domain characterized by sparse, delayed rewards and significant uncertainty. In this setting, we find that AIRL struggles to infer a sufficiently informative reward function. To overcome this limitation, we contribute Hybrid-AIRL (H-AIRL), an extension that enhances reward inference and policy learning by incorporating a supervised loss derived from expert data and a stochastic regularization mechanism. We evaluate H-AIRL on a carefully selected set of Gymnasium benchmarks and the HULHE poker setting. Additionally, we analyze the learned reward function through visualization to gain deeper insights into the learning process. Our experimental results show that H-AIRL achieves higher sample efficiency and more stable learning compared to AIRL. This highlights the benefits of incorporating supervised signals into inverse RL and establishes \mbox{H-AIRL} as a promising framework for tackling challenging, real-world settings.
\end{abstract}


\vspace{4ex}

\section{Introduction}\label{sec:introduction}
Deep reinforcement learning (RL) has recently demonstrated remarkable performance in complex tasks, including robotic control, epidemic control, and video games \cite{morales2021survey, mnih2015human, libin2021deep, oh2016control}. However, designing effective reward functions remains a critical challenge. In many settings, manually designed reward functions are often characterized by sparse or delayed feedback, which can hinder learning.

A compelling example of these challenges is found in strategic games, as their complex dynamics create a highly demanding learning environment. Poker, as a representative imperfect-information game, poses a particularly demanding scenario. In poker, the reward signal is inherently sparse, as feedback is generally available only at the conclusion of each hand, which restricts the information available for learning effective strategies. Historically, superhuman poker agents have been developed using Counterfactual Regret Minimization (CFR) \cite{neller2013introduction}, a method that iteratively converges toward a Nash equilibrium and guarantees optimal play. Despite its theoretical guarantees, CFR is computationally intensive and demands significant resources \cite{zhao2022alphaholdem}.

Reinforcement learning offers an attractive alternative by enabling agents to learn strategies directly from experience \cite{dahl2001reinforcement}. Yet, conventional RL approaches targeted at poker struggle with the inherent sparsity of the reward signal, which impedes the agent's ability to capture nuanced expert strategies. In this context, Inverse Reinforcement Learning (IRL) offers a promising approach by aiming to recover a dense reward function from expert demonstrations. Rather than merely imitating the expert, IRL seeks to uncover the underlying motivations driving expert behavior \cite{arora2021survey}.

Early IRL methods leveraged the maximum-entropy principle to model expert behavior, laying the foundation for probabilistic formulations that capture variability in demonstrations \cite{ziebart2008maximum}. Building on these ideas, adversarial formulations such as Adversarial Inverse Reinforcement Learning (AIRL) have extended the approach by jointly recovering the reward function and policy \cite{fu2018learning}. However, our investigation of AIRL in the context of poker reveals that AIRL has difficulty extracting a sufficiently informative reward function. This suggests that AIRL may struggle in domains that are characterized by large state-action spaces, high stochasticity, partial observability, and sparse rewards.

In response to these challenges, we contribute Hybrid Adversarial Inverse Reinforcement Learning (H-AIRL), a novel IRL method that leverages adversarial learning, supervised learning, and stochastic regularization. Inspired by techniques in conditional generative adversarial networks, our approach supplements the adversarial loss with an additional supervised signal, thereby improving sample efficiency and stabilizing training \cite{odena2017conditional}. Through systematic evaluations on carefully selected Gymnasium benchmarks and in Heads-Up Limit Hold'em (HULHE) poker, we demonstrate that H-AIRL exhibits improved sample efficiency and more stable learning compared to AIRL.

\section{Related Work}

Inverse Reinforcement Learning (IRL) is a framework for recovering the underlying reward function from expert demonstrations, thereby enabling an agent to infer the motivations behind expert behavior. Early work focused on linear reward models \cite{ng2000algorithms, abbeel2004apprenticeship}, and later probabilistic formulations recast the problem under the maximum-entropy principle \cite{ziebart2008maximum, boularias2011relative}. This approach establishes a probabilistic model over trajectories that captures the variability in expert behavior.

Building upon these ideas, adversarial methods for imitation learning have emerged. Generative Adversarial Imitation Learning (GAIL) leverages a discriminator to differentiate between expert and generated behavior, implicitly drawing on maximum-entropy formulations \cite{ho2016generative}. However, GAIL does not yield an explicit reward function. To overcome this limitation, Fu et al.\ introduced Adversarial Inverse Reinforcement Learning (AIRL), which reframes the discriminator as an odds ratio between the expert and policy behaviors to jointly infer a reward function and a policy \cite{fu2018learning}. More recent works such as Generative Intrinsic Reward-driven Imitation Learning (GIRIL) \cite{yu2020intrinsic} and Belief-Module Imitation Learning (BMIL) \cite{gangwani2020learning} have further addressed challenges like partial observability, though these approaches focus on imitation rather than explicitly recovering the latent reward function.

In the domain of poker, Counterfactual Regret Minimization (CFR) has long been the gold standard for developing superhuman agents by iteratively approximating Nash equilibria with strong theoretical guarantees \cite{li2021survey}. However, CFR-based methods are computationally intensive and require significant domain-specific tuning. Although reinforcement learning has emerged as a promising alternative for learning strategies directly from interactions, conventional RL methods struggle with sparse, stochastic, and delayed rewards in poker \cite{zhao2022alphaholdem}.

IRL offers the potential to mitigate these challenges by inferring dense reward signals from expert demonstrations, thus providing richer guidance than the sparse, terminal rewards typical of RL. Despite the maturity of RL methods in poker, the application of inverse RL to capture the nuanced reward incentives in this domain has not yet been explored. Our experimental results reveal that AIRL struggles to extract sufficiently informative reward functions in HULHE poker, underscoring its limitations in such complex environments.

To address these gaps, our work introduces a novel hybrid IRL framework, H-AIRL, which incorporates a supervised loss component derived directly from expert data. While supervised signals have been successfully integrated into generative adversarial networks for tasks such as conditional image synthesis \cite{odena2017conditional}, no prior work has fused a similar supervised term within an IRL framework. By leveraging this additional guidance, H-AIRL aims to stabilize training and enhance the quality of the recovered reward function, particularly in challenging environments like poker.

In summary, while significant progress has been made in both imitation learning and RL for complex tasks, our work is the first to integrate a supervised loss component into an IRL framework and to apply IRL to the challenging domain of poker.

\section{Background}\label{sec:background}

\subsection{Reinforcement Learning}
Reinforcement Learning (RL) is one of three main machine learning paradigms alongside supervised and unsupervised learning \cite{mitchell1997machine}. In RL, an agent interacts with an environment by taking actions, receiving feedback in the form of rewards, and adjusting its behavior to maximize cumulative expected returns \cite{mitchell1997machine, sutton2018reinforcement}.

Formally, RL problems are modeled as Markov Decision Processes (MDPs). An MDP is defined by the tuple $\langle \mathcal{S}, \mathcal{A}, P, r, \gamma \rangle$, where $\mathcal{S}$ is the set of states, $\mathcal{A}$ is the set of actions, $P(\mathbf{s}'|\mathbf{s},\mathbf{a})$ describes the environment's transition dynamics, $r(\mathbf{s},\mathbf{a})$ is the reward function, and $\gamma \in [0,1]$ is a discount factor. At each time step, the agent observes a state $\mathbf{s} \in \mathcal{S}$, selects an action $\mathbf{a} \in \mathcal{A}$, after which it transitions to a new state $\mathbf{s}'$ while receiving a scalar reward $r(\mathbf{s},\mathbf{a})$. The goal is to find a policy $\pi(\mathbf{a}|\mathbf{s})$ that maximizes the expected discounted return.

Q-learning is a foundational RL algorithm that learns an action-value function $Q(\mathbf{s},\mathbf{a})$, representing the expected return for taking action $\mathbf{a}$ in state $\mathbf{s}$ and following the optimal policy thereafter \cite{watkins1992q}. In the tabular setting, Q-learning is proven to converge to the optimal policy under mild conditions \cite{sutton2018reinforcement}. In high-dimensional or continuous state spaces, function approximators such as deep neural networks can be employed to estimate $Q(\mathbf{s},\mathbf{a})$. This approach led to the development of Deep Q-Networks (DQN), an algorithm that achieved human-level performance on Atari games \cite{mnih2015human}. Despite these successes, a key assumption in RL is that the reward function $r(\mathbf{s},\mathbf{a})$ is known or specified by the designer, a constraint that motivates the field of Inverse Reinforcement Learning.

\subsection{Inverse Reinforcement Learning}
Inverse Reinforcement Learning (IRL) inverts the classical RL paradigm by inferring the underlying reward function from expert demonstrations rather than assuming it is given \cite{russell1998learning, ng2000algorithms}. Formally, consider an MDP defined by the tuple $\langle \mathcal{S}, \mathcal{A}, P, r, \gamma \rangle$. In standard RL one seeks an optimal policy $\pi^*(\mathbf{a}|\mathbf{s})$ that maximizes the cumulative reward. In IRL, the agent observes expert demonstrations, consisting of trajectories sampled from a policy $\pi^*(\mathbf{a}|\mathbf{s})$ that is assumed to be optimal, and aims to infer a reward function $f(\mathbf{s},\mathbf{a})$ that explains the expert's behavior.

Early IRL algorithms, which focused on linear reward models, revealed the fundamental \textit{degeneracy} in IRL, namely, that multiple distinct reward functions can yield the same optimal policy \cite{ng2000algorithms, abbeel2004apprenticeship}. To mitigate this ambiguity, Ziebart et al.\ introduced the \textit{maximum-entropy} IRL framework, which assigns a probability to each trajectory that both explains the expert's behavior and favors high-entropy policies \cite{ziebart2008maximum}. In this formulation, the expert's trajectory distribution $\rho_\theta(\tau)$ is given by:
$$
\rho_\theta(\tau) \;\propto\; \exp\!\Biggl(\sum_{t=0}^{T} r_\theta(\mathbf{s}_t,\mathbf{a}_t)\Biggr),
$$

where $\tau = (\mathbf{s}_0, \mathbf{a}_0, \ldots, \mathbf{s}_T, \mathbf{a}_T)$ denotes a trajectory and $r_\theta(\mathbf{s},\mathbf{a})$ is a reward function with parameters $\theta$. To transform this expression into a valid probability distribution, a normalizing constant known as the partition function is required:
$$
Z_\theta = \sum_{\tau} \exp\!\Biggl(\sum_{t=0}^{T} r_\theta(\mathbf{s}_t,\mathbf{a}_t)\Biggr).
$$

In high-dimensional spaces, calculating $Z_\theta$ is typically intractable. Methods such as Guided Cost Learning (GCL) address this issue by using function approximators (e.g., neural networks) to estimate the reward function without requiring explicit computation of $Z_\theta$ \cite{finn2016guided}. Nonetheless, scaling maximum-entropy IRL to complex, high-dimensional problems remains challenging \cite{fu2018learning}. This motivated the development of alternative approaches. 

In particular, Generative Adversarial Imitation Learning (GAIL) represents a significant milestone in imitation learning \cite{ho2016generative}. GAIL reframes imitation as a min-max game between a generator (the policy) and a discriminator. The discriminator $D_\phi(\mathbf{s},\mathbf{a})$ is trained to distinguish expert state-action pairs from those generated by the policy $\pi(\mathbf{a}|\mathbf{s})$, while the policy is optimized to deceive the discriminator. Although GAIL is not strictly an IRL method, as it does not recover an explicit reward function, it represents a significant milestone in imitation learning and has inspired subsequent work in IRL. Notably, its adversarial formulation paved the way for Adversarial Inverse Reinforcement Learning (AIRL) \cite{fu2018learning}.

\subsubsection{The Adversarial IRL Framework.}  
The AIRL algorithm expands on GAIL by integrating the maximum-entropy formulation of IRL into an adversarial framework. AIRL jointly learns a policy and a reward function by reinterpreting the discriminator as an energy-based model. In AIRL, the discriminator is defined as:
\begin{equation}
    D_\theta(\mathbf{s},\mathbf{a}) \;=\; \frac{\exp\bigl(f_\theta(\mathbf{s},\mathbf{a})\bigr)}{\exp\bigl(f_\theta(\mathbf{s},\mathbf{a})\bigr) + \pi_\phi(\mathbf{a}|\mathbf{s})},
    \label{eq:background:irl:airl:discriminator}
\end{equation}

where $\pi_\phi(\mathbf{a}|\mathbf{s})$ denotes the probability of taking action $\mathbf{a}$ in state $\mathbf{s}$ under policy $\pi_\phi$ with parameters $\phi$, and $f_\theta(\mathbf{s},\mathbf{a})$ is the learned reward function with parameters $\theta$. Meanwhile, the policy objective is defined as:
\begin{equation}
    \max_{\phi}\;\mathbb{E}_{\tau \sim \pi_\phi}\!\Biggl[\sum_{t=0}^{T}\Bigl(f_\theta(\mathbf{s}_t,\mathbf{a}_t) - \log \pi_\phi(\mathbf{a}_t|\mathbf{s}_t)\Bigr)\Biggr].
    \label{eq:background:irl:airl:objective}
\end{equation}

From an optimization perspective, maximizing the policy objective is equivalent to minimizing a loss function defined as its negative:

\begin{equation}
    \mathcal{L}^{\text{policy}}_{\text{AIRL}} = -f_\theta(\mathbf{s},\mathbf{a}) + \log \pi_\phi(\mathbf{a}|\mathbf{s}).
    \label{eq:background:irl:airl:loss}
\end{equation}

Here, the negated reward term $-f_\theta(\mathbf{s},\mathbf{a})$ encourages the policy to favor actions that yield higher rewards from the learned reward function. Meanwhile, the entropy regularization term $\log \pi_\phi(\mathbf{a}|\mathbf{s})$ helps maintain exploration by increasing the loss for actions to which the policy assigns a high probability.

So far, we have described AIRL in its state-action form. The \textit{state-only} variant constrains the learned reward to:
$$
f_{\theta,\omega}(\mathbf{s},\mathbf{a},\mathbf{s}')
= g_{\theta}(\mathbf{s},\mathbf{a})\;+\;\gamma\,h_{\omega}(\mathbf{s}')
\;-\;h_{\omega}(\mathbf{s}),
$$
where $g_{\theta}(\mathbf{s},\mathbf{a})$ captures the true reward (up to an additive constant) and $h_{\omega}(\mathbf{s})$ is a potential‐based shaping term. In this decomposition, $g_{\theta}(\mathbf{s},\mathbf{a})$ alone is a disentangled reward function that remains valid when transferred to new MDPs with different transition dynamics. Crucially, however, learning the state-only form requires access to the actual next state $\mathbf{s}'$, which is infeasible in many offline or partially observable domains where a faithful simulation model is not available. In such cases, one must revert to the unrestricted state-action form. Accordingly, in the remainder of this paper, we adopt AIRL's state-action formulation as our baseline.

\section{The Hybrid-AIRL Framework}\label{sec:h-airl}
The Hybrid-AIRL algorithm introduces a novel integration of supervised learning into both the policy and the discriminator within the adversarial maximum-entropy framework.

\subsection{The Policy Objective}
Under the maximum-entropy principle, the probability of a trajectory $\tau$ can be modeled as:
$$
\rho_\theta(\tau) =p(\mathbf{s}_0) \prod_{t=0}^{T-1} p(\mathbf{s}_{t+1} \mid \mathbf{s}_t,\mathbf{a}_t) \; \exp\!\Biggl(\sum_{t=0}^{T} f_\theta(\mathbf{s}_t,\mathbf{a}_t)\Biggr) \frac{1}{Z_\theta},
$$

where $p(\mathbf{s}_0)$ is the probability of the initial state, $p(\mathbf{s}_{t+1} \mid \mathbf{s}_t,\mathbf{a}_t)$ represents the environment dynamics, and $Z_\theta$ is the partition function ensuring normalization. As such, $\rho_\theta(\tau)$ represents the probability distribution over trajectories given reward function $f_\theta(\mathbf{s},\mathbf{a})$, reflecting which trajectories are desirable based on the model's current parametrization $\theta$ of the reward function. Taking the logarithm yields:
\begin{align*}
\log \rho_\theta(\tau) &= \sum_{t=0}^{T} f_\theta(\mathbf{s}_t,\mathbf{a}_t) - \log Z_\theta \\
                       &\quad + \log p(\mathbf{s}_0) + \sum_{t=0}^{T-1} \log p(\mathbf{s}_{t+1} \mid \mathbf{s}_t,\mathbf{a}_t).
\end{align*}

We then define the IRL objective $J(\theta)$ as the expected log-likelihood:
$$
J(\theta) = \mathbb{E}_{\tau \sim \rho_\theta}\!\Bigl[\log \rho_\theta(\tau)\Bigr].
$$

In maximum-entropy IRL frameworks such as state-action AIRL, the goal is to align the policy's trajectory distribution $\pi_\phi$ with the target distribution $\rho_\theta$, which represents the probability of trajectories based on the current parametrization $\theta$ of the reward function. This can be enforced by minimizing the KL divergence: 
\begin{equation}
    \mathcal{D}_{\text{KL}}\bigl(\pi_\phi \,\|\, \rho_\theta\bigr) = \mathbb{E}_{\tau \sim \pi_\phi}\!\Bigl[\log \frac{\pi_\phi(\tau)}{\rho_\theta(\tau)}\Bigr].
    \label{eq:h-airl:kl_divergence_airl}
\end{equation}

Maximizing the negative of this divergence leads to the objective:
\begin{equation}
\max_{\phi}\;\mathbb{E}_{\tau \sim \pi_\phi}\!\Bigl[\log \rho_\theta(\tau) - \log \pi_\phi(\tau)\Bigr].
    \label{eq:h-airl:max_negative_divergence}
\end{equation}

Because the environment is assumed to satisfy the Markov property, the probability of a trajectory $\tau$ under the policy $\pi_\phi$ factorizes as:
$$
\pi_\phi(\tau) = p(\mathbf{s}_0)\; \prod_{t=0}^{T-1} p(\mathbf{s}_{t+1} \mid \mathbf{s}_t,\mathbf{a}_t)\, \pi_\phi(\mathbf{a}_t|\mathbf{s}_t).
$$

Since both $\rho_\theta(\tau)$ and $\pi_\phi(\tau)$ include the initial state probability $p(\mathbf{s}_0)$ and the environment dynamics $p(\mathbf{s}_{t+1} \mid \mathbf{s}_t,\mathbf{a}_t)$ in their definitions, these terms cancel out when computing the difference:
\begin{equation}
    \log \rho_\theta(\tau) - \log \pi_\phi(\tau) \propto \sum_{t=0}^{T}\Bigl(f_\theta(\mathbf{s}_t,\mathbf{a}_t) - \log \pi_\phi(\mathbf{a}_t|\mathbf{s}_t)\Bigr).
    \label{eq:h-airl:log_distribution_difference}
\end{equation}

Finally, using Equations \eqref{eq:h-airl:max_negative_divergence} and \eqref{eq:h-airl:log_distribution_difference}, we find that the policy objective is equivalent to maximizing the expected cumulative entropy-regularized reward:
\begin{equation*}
    \max_{\phi}\;\mathbb{E}_{\tau \sim \pi_\phi}\!\Biggl[\sum_{t=0}^{T}\Bigl(f_\theta(\mathbf{s}_t,\mathbf{a}_t) - \log \pi_\phi(\mathbf{a}_t|\mathbf{s}_t)\Bigr)\Biggr].
\end{equation*}

This expression precisely matches the policy objective of AIRL, described by Equation \eqref{eq:background:irl:airl:objective}. In essence, this objective is derived from the standard maximum-entropy IRL framework, which is solely concerned with aligning the policy's trajectory distribution with that of the expert.

In our H-AIRL framework, we extend this objective by aligning the policy's \textit{action} distribution with the expert's. That is, in addition to matching trajectories, we incorporate a supervised learning objective that minimizes the discrepancy between the policy's actions and the expert's actions. We define this hybrid policy objective as:
\begin{equation}
    \min_{\phi} \Biggl[ (1\!-\!\alpha) \!\!\underbrace{\mathcal{D}_{\text{KL}} \! \bigl( \pi_\phi \| \rho_\theta \bigr)}_{\substack{\text{(1) Max-Entropy IRL}\\\text{Objective}}} + \alpha \underbrace{\mathbb{E}_{\tau\sim\rho_E}\!\Bigl[-\!\sum_{t=0}^{T}\log\pi_\phi(\mathbf{a}_t|\mathbf{s}_t)\Bigr]}_{\text{(2) Supervised Learning Objective}}\Biggr], 
    \label{eq:h-airl:hybrid_formulation}
\end{equation}

where $\rho_E$ is the distribution of expert trajectories and $\alpha \in [0,1]$ is a weighting factor. The first term in Equation \eqref{eq:h-airl:hybrid_formulation} coincides with the maximum-entropy IRL objective as defined in Equation \eqref{eq:h-airl:kl_divergence_airl}, which leads to AIRL's policy loss formulation in Equation \eqref{eq:background:irl:airl:loss}. The second term constitutes the supervised learning objective. Consequently, the policy's loss function becomes:
\begin{align*}
\mathcal{L}^{\text{policy}}_{\text{H-AIRL}} &= (1-\alpha) \mathcal{L}^{\text{policy}}_{\text{AIRL}} + \alpha\,\mathcal{L}^{\text{policy}}_{\text{S}} \\
&= (1-\alpha) \Bigl[-f_\theta(\mathbf{s},\mathbf{a}) + \log \pi_\phi(\mathbf{a}|\mathbf{s})\Bigr] + \alpha\,\mathcal{L}^{\text{policy}}_{\text{S}} \;,
\end{align*}

where $\mathcal{L}_{\text{S}}$ denotes the supervised loss component. This hybrid objective benefits from both adversarial IRL and direct supervised imitation. For discrete action spaces, the supervised loss can be computed as the Cross-Entropy loss or KL divergence between the policy's action distribution $\pi_\phi(\cdot|\mathbf{s})$ and the expert's action distribution $\rho_E(\cdot|\mathbf{s})$. For continuous action spaces, this can be approximated using an appropriate loss function (e.g., the mean-squared-error loss). 

\subsection{The Discriminator Objective}\label{sec:h-airl:disc}
Given the definition of the discriminator in Equation \eqref{eq:background:irl:airl:discriminator} and a fixed policy $\pi_\phi$, AIRL optimizes the discriminator parameters $\theta$ via the standard cross-entropy loss:

\begin{equation}
\begin{split}
\mathcal{L}^{\text{disc}}_{\text{AIRL}}
&= -\mathbb{E}_{(\mathbf{s},\mathbf{a})\sim\rho_E}\!\bigl[\log D_\theta(\mathbf{s},\mathbf{a})\bigr] \\
&\quad -\mathbb{E}_{(\mathbf{s},\mathbf{a})\sim\pi_\phi}\!\bigl[\log\bigl(1 - D_\theta(\mathbf{s},\mathbf{a})\bigr)\bigr].
\end{split}
\label{eq:h-airl:disc_airl}
\end{equation}

This adversarial objective encourages $f_\theta$ to assign higher scores to expert
state-action pairs, thereby inferring a reward function that explains the demonstrations.

In AIRL, the discriminator learns to distinguish expert from policy-generated trajectories. However, adversarial training alone provides no guarantee that the learned reward $f_\theta$ aligns with the true environment reward $r_{\text{env}}$. Optimizing a policy under $f_\theta$ thus defines a shaped MDP, which has the same $(\mathcal S,\mathcal A,P,\gamma)$ but with reward $f_\theta$ in place of $r_{\text{env}}$. A policy that performs well in this shaped MDP may perform poorly when evaluated under the original reward \cite{ng1999policy, wiewiora2003principled}. To prevent this deviation, we regularize $f_\theta$ with an additional supervised mean‐squared‐error loss when ground‐truth environment rewards $r_{\text{env}}(\mathbf{s},\mathbf{a})$ are available for the expert demonstrations:

\begin{equation}
\mathcal{L}^{\text{disc}}_{\text{S}}
  = \mathbb{E}_{(\mathbf{s},\mathbf{a})\sim\rho_E}
      \bigl[\,(f_\theta(\mathbf{s},\mathbf{a}) - r_{\text{env}}(\mathbf{s},\mathbf{a}))^{2}\bigr].
\label{eq:h-airl:disc_gt}
\end{equation}

By combining the adversarial and supervised loss components, we obtain a new discriminator objective:

\begin{equation*}
\mathcal{L}^{\text{disc}}_{\text{AIRL+S}} = (1-\beta)\mathcal{L}^{\text{disc}}_{\text{AIRL}} + \beta\,\mathcal{L}^{\text{disc}}_{\text{S}},
\label{eq:h-airl:airl+s}
\end{equation*}

where $\beta \in [0,1]$ is a weighting factor that balances the adversarial loss against supervised loss. This approach aims to prevent pathological reward shaping and to ensure that a policy trained using the learned $f_\theta$ continues to perform well when evaluated in the original MDP. If ground-truth rewards are unavailable, we set $\beta=0$.

\subsection{Stochastic Regularization}\label{sec:h-airl:noise}
In the hybrid policy objective (Equation \ref{eq:h-airl:hybrid_formulation}), the supervised term can rapidly drive the policy $\pi_\phi$ toward expert-like outputs, as demonstrated by our experimental results in Section \ref{sec:results}. Consequently, the discriminator $D_\theta$ is exposed primarily to high-quality actions and receives limited feedback for distinguishing realistic from unrealistic behavior, which may lead to overfitting.

To mitigate this effect and restore a meaningful adversarial signal, we introduce stochastic regularization by injecting Gaussian noise that decays along the mini-batch axis. Let $\mathcal B = \{(\mathbf{s}_i,\mathbf{a}_i)\}_{i=0}^{B-1}$ represent a mini-batch of size $B$ that contains state-action pairs $(\mathbf{s}_i,\mathbf{a}_i)$ associated with the policy $\pi_\phi$. For every index $i$ in the batch, we form a perturbed action vector $\tilde{\mathbf{a}}_i$ using Gaussian noise:
$$
\tilde{\mathbf{a}}_i = \mathbf{a}_i + \boldsymbol{\eta}_i,
\qquad
\boldsymbol{\eta}_i \sim \mathcal N\!\bigl(\mathbf 0,\;\sigma_i^{2}\mathbf I\bigr),
$$
where $\sigma_i$ is the standard deviation and $\tilde{\mathbf{a}}_i$ is the perturbed policy action. The standard deviation per sample $\sigma_i$ decays monotonically from the first element of the mini-batch to the last. As a result, each batch that reaches the discriminator $D_\theta$ contains a spectrum of action qualities, ranging from strongly perturbed (large $\sigma_i$) to nearly noise-free (small $\sigma_i$).

By combining these regularized adversarial and supervised losses, we obtain the Hybrid-AIRL discriminator objective:
\begin{equation}
\mathcal{L}^{\text{disc}}_{\text{H-AIRL}} = (1-\beta)\mathcal{L}^{\text{disc}}_{\text{AIRL+SR}} + \beta\,\mathcal{L}^{\text{disc}}_{\text{S+SR}}.
\label{eq:h-airl:disc_h-airl}
\end{equation}

\section{Experimental Setup}
The IRL training procedure of H-AIRL\footnote{\url{https://github.com/silue-dev/hairl}} alternates between:

\begin{enumerate}
    \item Training the discriminator to classify expert trajectories from those generated by the policy.
    \item Updating the policy to maximize the inferred reward.
\end{enumerate}

Once this IRL training process completes, the discriminator effectively becomes a surrogate for the expert's reward function, encapsulating the underlying reward incentives observed in the expert demonstrations. This learned reward function serves to both capture the underlying motivations of expert behavior and provide a dense signal for reinforcement learning. Thus, we subsequently consider an RL training phase, where we integrate this learned reward function into an RL agent, namely, a Proximal Policy Optimization (PPO) agent from Stable-Baselines3 for Gymnasium benchmarks and a DQN agent from the RLCard framework for HULHE poker \cite{raffin2021stable, zha2020rlcard}. 

\subsection{Benchmarks}\label{sec:methods:the_data}
We first benchmark AIRL and H-AIRL on a selection of Gymnasium tasks: Pendulum, Ant, HalfCheetah, Acrobot, LunarLander, and MountainCar \cite{towers2024gymnasium}. The first three tasks (Pendulum, Ant, and Half‑Cheetah) are the same continuous‑control benchmarks that were used in the original AIRL paper, making them well-suited for comparison with our approach \cite{fu2018learning}. On the other hand, Acrobot, LunarLander, and MountainCar are discrete environments, which complement the continuous set and broaden the scope of our evaluation. To obtain expert data for these tasks, we train a PPO agent \cite{schulman2017proximal}, following the approach used in AIRL.\footnote{Fu et al.\ employed TRPO rather than PPO, its successor \cite{fu2018learning, schulman2015trust}. For MountainCar, we exceptionally use DQN \cite{mnih2015human}.}

Next, we explore a more challenging task: HULHE poker. This imperfect‑information, zero‑sum game features partial observability, stochastic dynamics, and a vast state-action space. Rewards are only observed at the end of each hand, which makes strategic learning particularly difficult. For this task, we use the IRC Poker dataset\footnote{\url{https://poker.cs.ualberta.ca/irc_poker_database.html}}, a high‑quality collection of online poker games featuring both expert amateur players and professional players, including a world champion. This extensive dataset comprises millions of real-world poker game state observations, including over 1 million state-action pairs for HULHE poker.

We note that a distinct challenge in online poker datasets is that folding actions reveal no information about the player's hand, as folded cards remain concealed and are therefore absent from the data. This limitation poses a significant obstacle for machine learning algorithms, particularly IRL agents, which rely on observable behavior to infer the underlying reward function. Without the ability to learn from folding actions, the IRL agent must focus on mastering the remaining actions (i.e., calling, raising, and checking) to infer the expert strategy.

\subsection{Evaluation}\label{sec:methods:evaluation}
To evaluate the effectiveness of AIRL and H-AIRL, we assess both the alignment of the learned policies with expert behavior and the quality of the inferred reward functions.

First, for Gymnasium benchmarks, we assess the learning curves of the reward obtained by the agent during training. This widely used reinforcement learning evaluation method offers insight into the agent's sample efficiency and performance under the inferred reward function.

For HULHE poker, where we consider a real-world dataset, we introduce the \textit{state‑level action alignment} as a complementary metric to assess IRL training performance. We define this metric as the percentage of visited states in which the learned policy selects the same action as the expert. This metric provides insight into the model's ability to mimic the expert's probabilistic decision‑making tendencies. Because this metric is applicable to any discrete action space environment, we also compute it for Gymnasium benchmarks with discrete action spaces.
 
Next, we extend our evaluation beyond the policy and assess the quality of the learned reward function. One key limitation of prior IRL work is the lack of direct validation of the inferred reward signal. To address this, we train separate RL agents (i.e., PPO or DQN) using the learned reward function and analyze their learning curves with respect to the environment reward. If the inferred reward function accurately captures task-relevant features, agents should be able to learn effective policies using only the IRL-derived reward. For poker, we integrate the learned reward function into the DQN implementation provided by RLCard, an open‑source library for applying reinforcement learning algorithms to card game environments \cite{zha2020rlcard}. Note that for all learning curves, we depict the mean and standard deviation across 10 training runs for each algorithm \cite{agarwal2021deep}.

For poker, reward learning curves during the RL phase can be less informative, as the agent is trained against a random opponent — an adversary against which even simplistic strategies, such as always raising, can achieve positive expected returns. Consequently, we further investigate the effectiveness of IRL-derived rewards in poker by directly opposing DQN agents trained with the learned dense reward versus those trained with the traditional sparse reward (game payoff). This allows us to assess whether these IRL models can compete with standard RL approaches. To ensure a robust comparison, we use the RLCard framework to simulate 1,000,000 tournaments across 20 random seeds \cite{zha2020rlcard}. Performance is measured in milli-big-blinds per hand (mbb/h), a standard poker metric reflecting average gains or losses per hand, normalized across games.

Finally, for MountainCar, an environment with a two-dimensional state space and three discrete actions, we generate directly interpretable 2D visualizations for both AIRL and H-AIRL. These visualizations illustrate the reward function's preferred action at each state.

\section{Experimental Results}\label{sec:results}

\subsection{IRL Training}\label{sec:results:irl}
\begin{figure}[H]
    \centering
    \begin{subfigure}{0.32\textwidth}
        \centering
        \includegraphics[width=\linewidth]{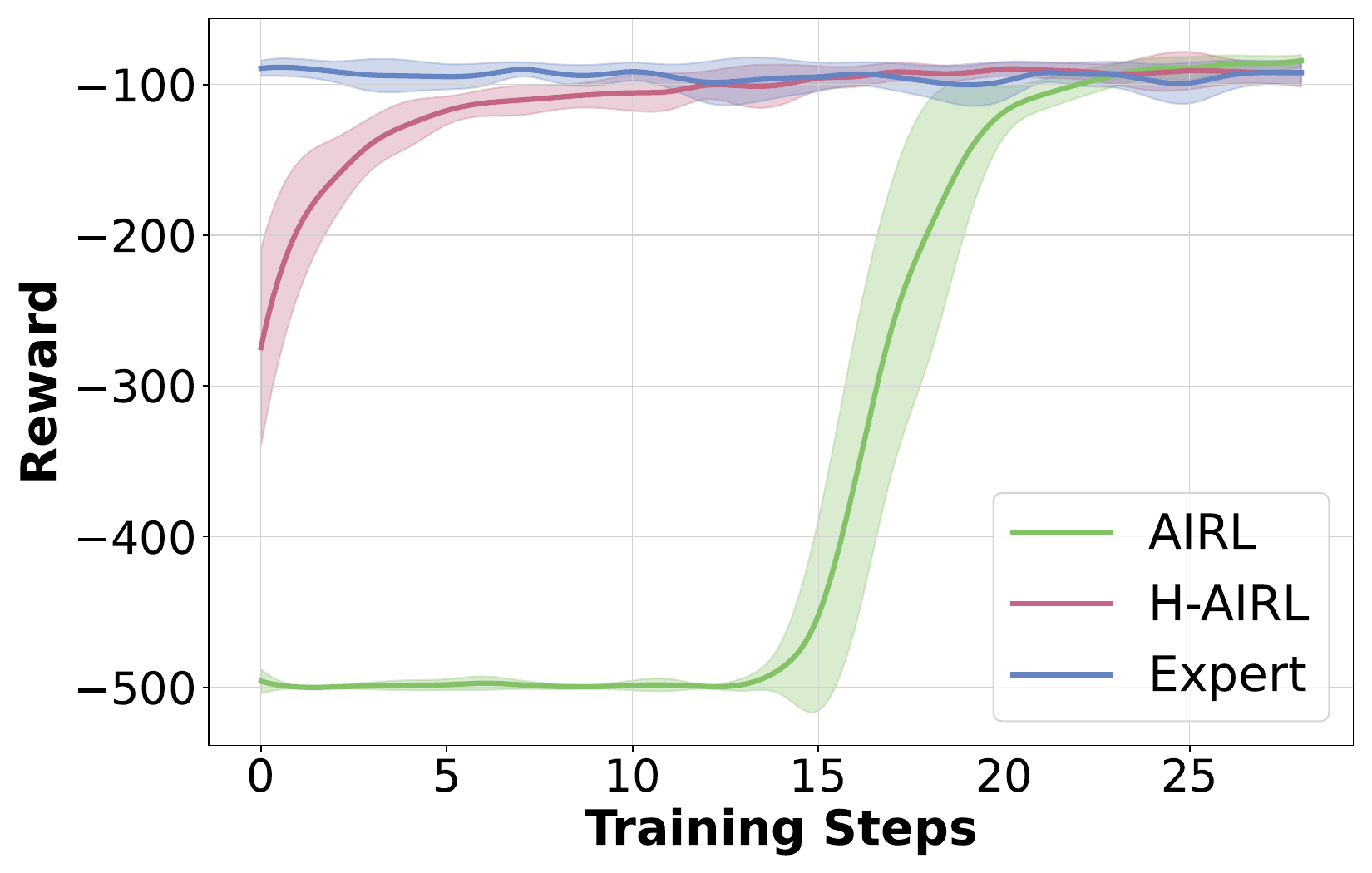}
        \caption{\small Acrobot}
        \label{fig:results:irl:rewards:acrobot}
    \end{subfigure}
    \hfill
    \begin{subfigure}{0.32\textwidth}
        \centering
        \includegraphics[width=\linewidth]{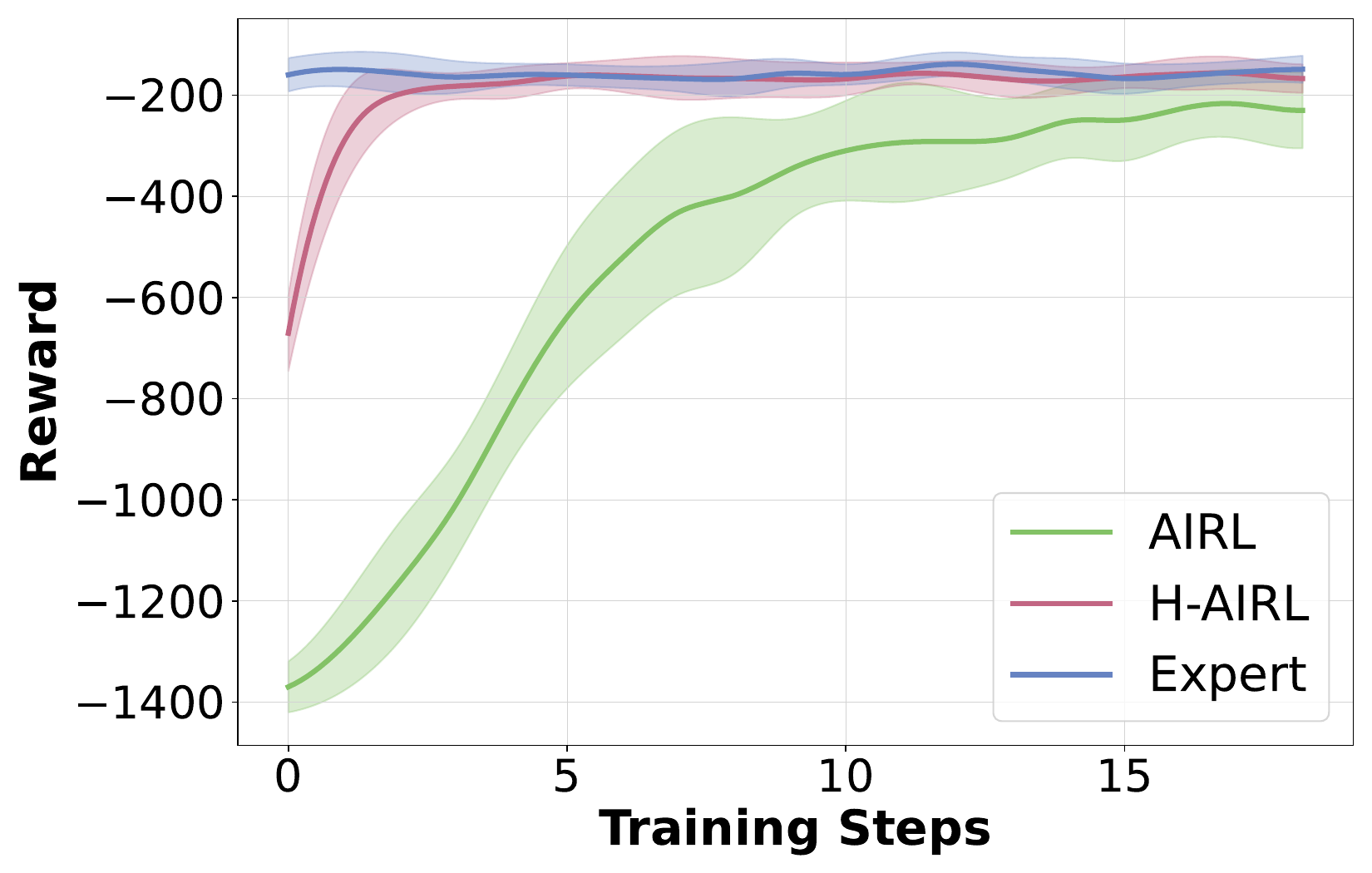}
        \caption{\small Pendulum}
        \label{fig:results:irl:rewards:pendulum}
    \end{subfigure}
    \hfill
    \begin{subfigure}{0.32\textwidth}
        \centering
        \includegraphics[width=\linewidth]{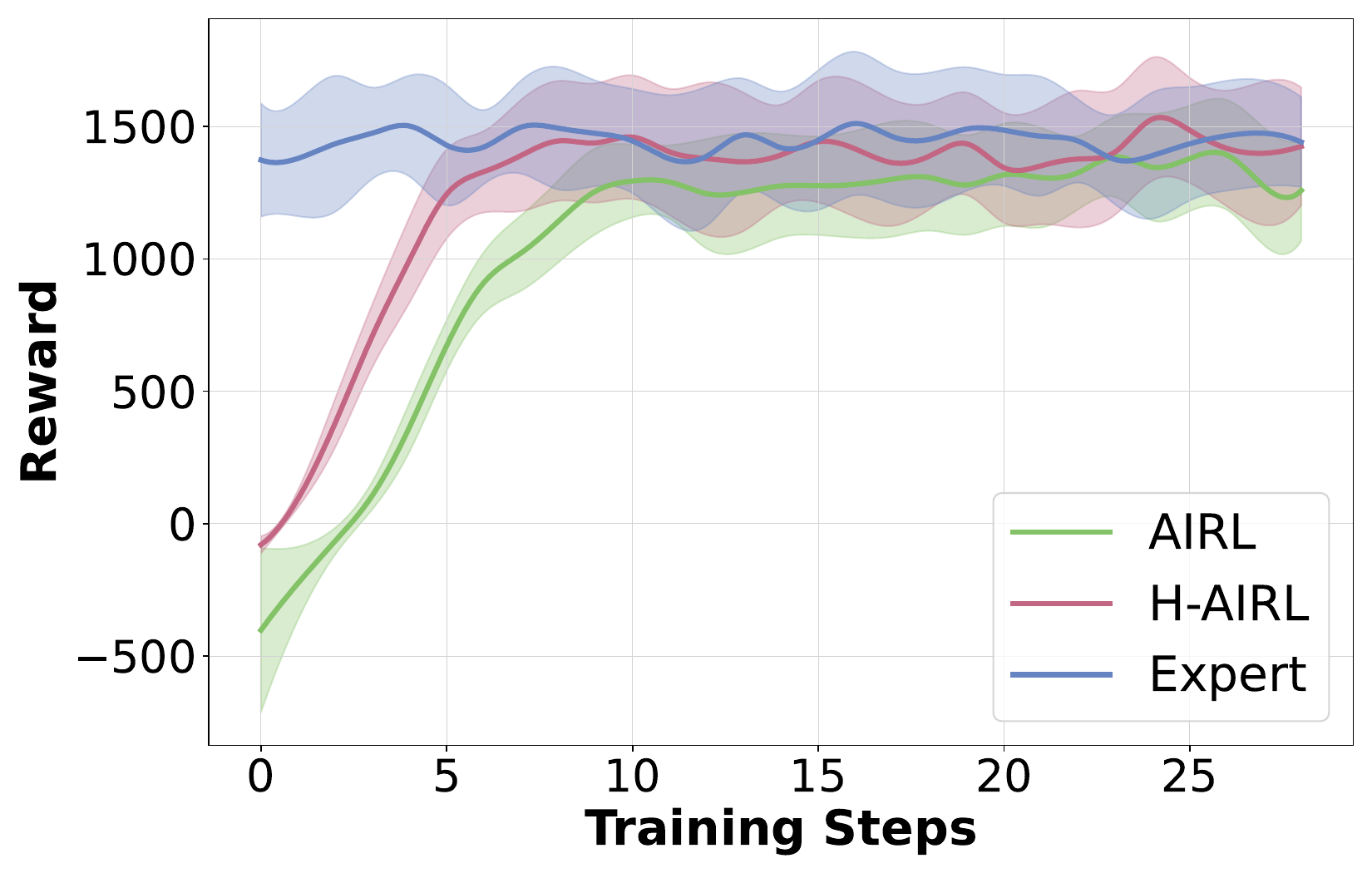}
        \caption{\small Ant}
        \label{fig:results:irl:rewards:ant}
    \end{subfigure}

    \vspace{2ex}

    \begin{subfigure}{0.32\textwidth}
        \centering
        \includegraphics[width=\linewidth]{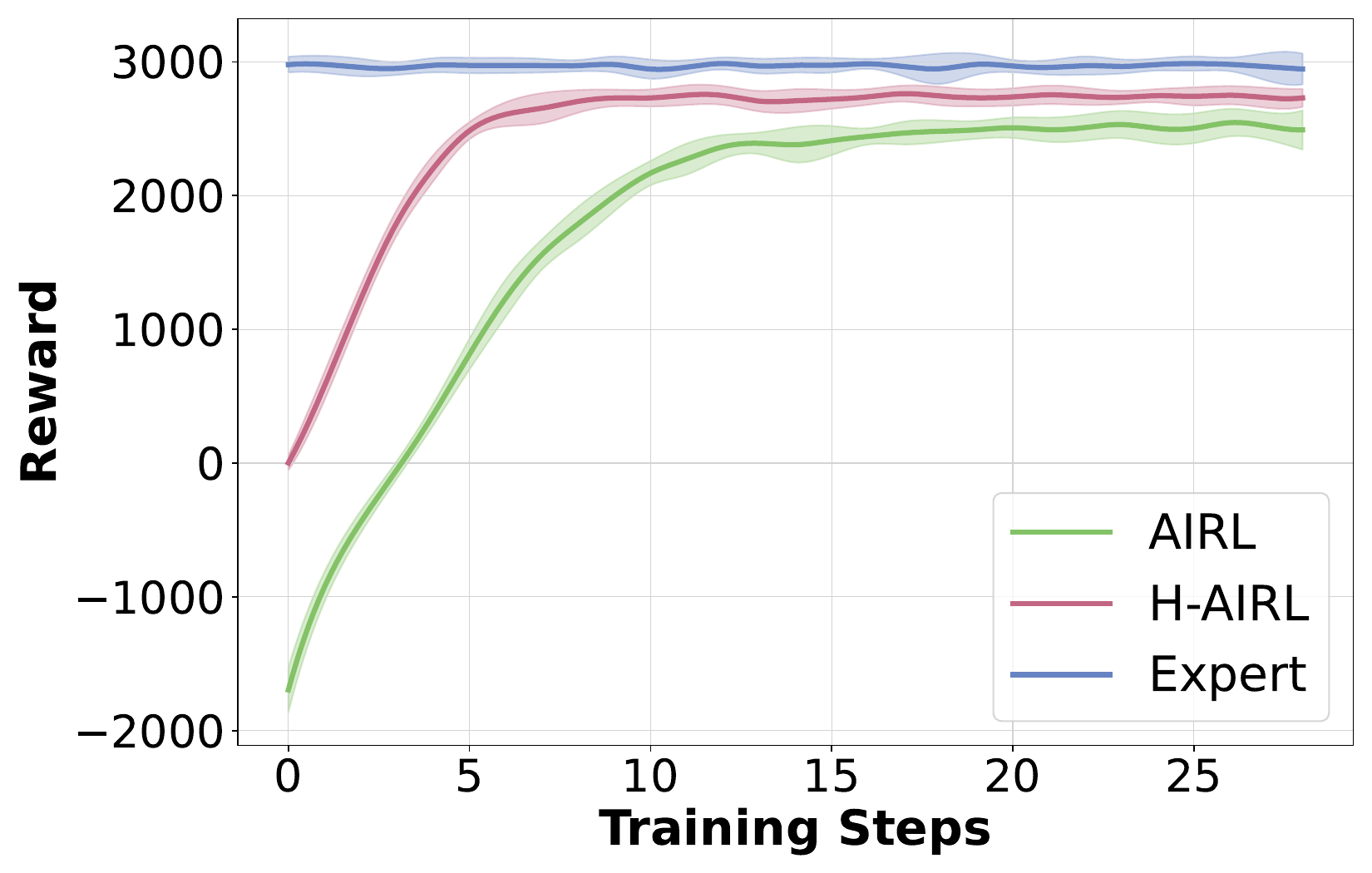}
        \caption{\small HalfCheetah}
        \label{fig:results:irl:rewards:halfcheetah}
    \end{subfigure}
    \hfill
    \begin{subfigure}{0.32\textwidth}
        \centering
        \includegraphics[width=\linewidth]{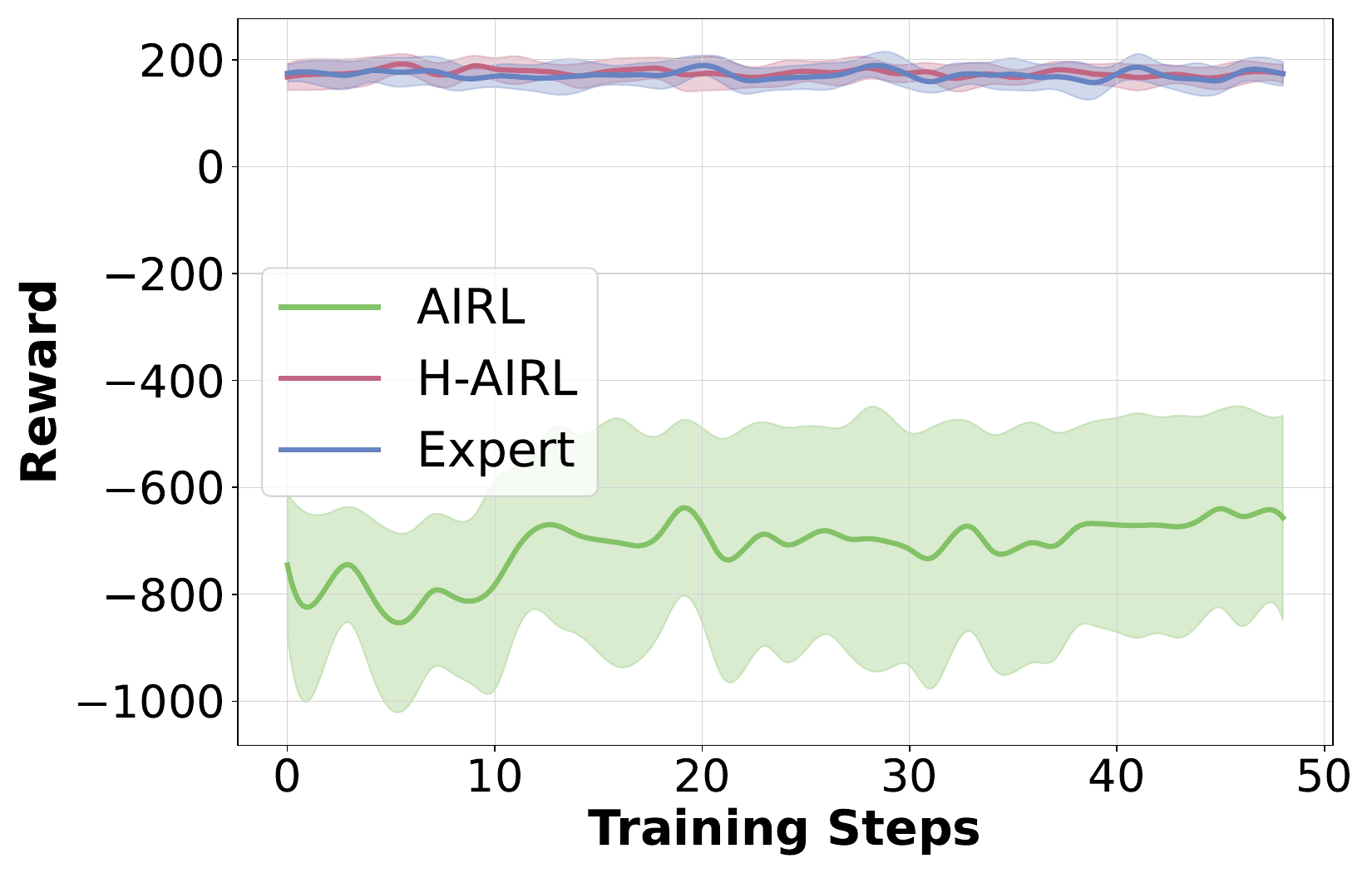}
        \caption{\small LunarLander}
    \end{subfigure}
    \hfill
    \begin{subfigure}{0.32\textwidth}
        \centering
        \includegraphics[width=\linewidth]{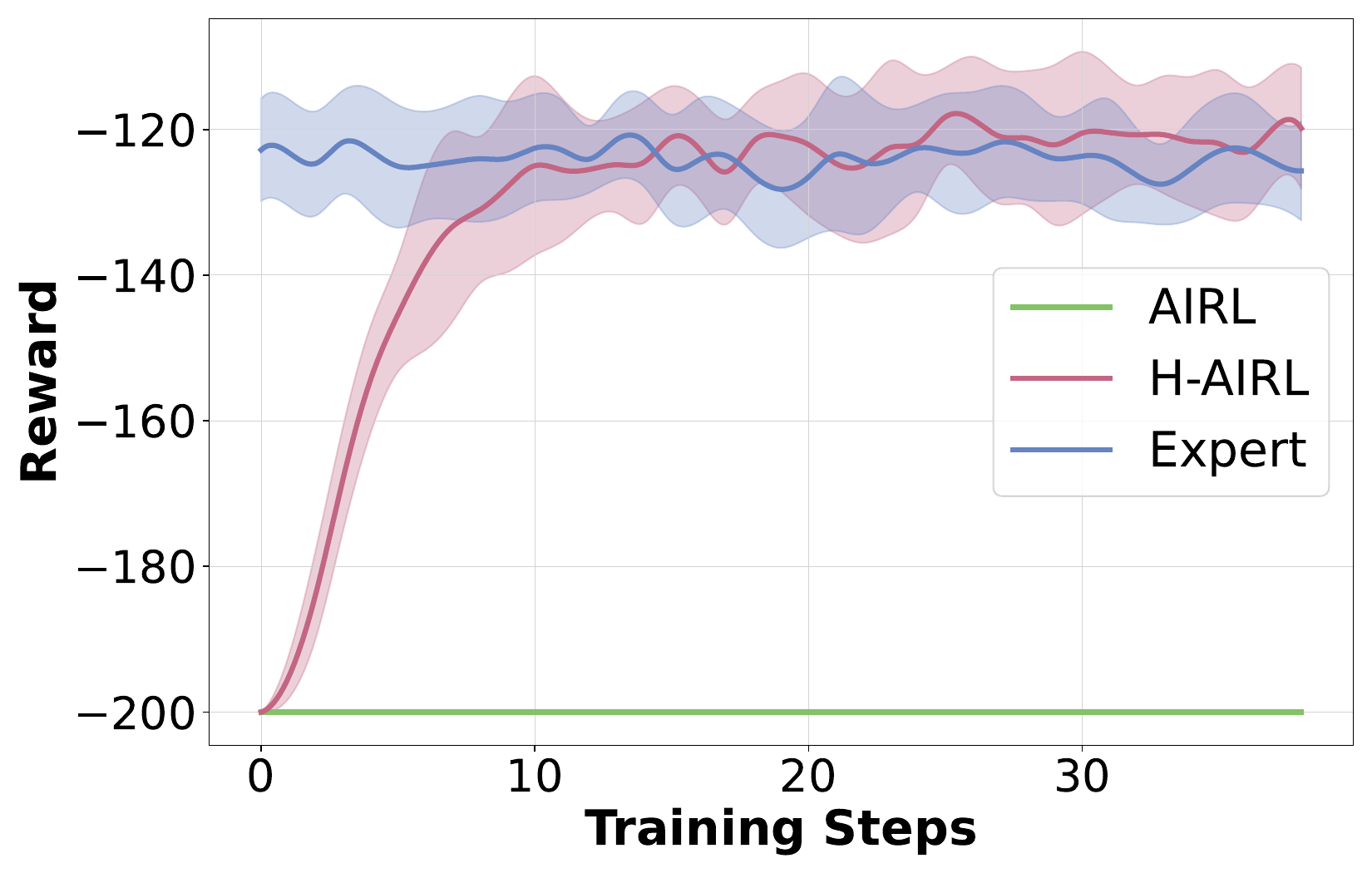}
        \caption{\small MountainCar}
    \end{subfigure}

    \captionsetup{width=0.90\textwidth}
    \caption{\small Reward learning curves for AIRL (green) and H-AIRL (red) on Gymnasium benchmarks, alongside an expert PPO baseline (blue).}
    \label{fig:results:irl:rewards}
\end{figure}

We present the performance of AIRL and H-AIRL during the IRL training phase. Figure \ref{fig:results:irl:rewards} shows the reward learning curves on the Gymnasium benchmarks.

\begin{figure}[H]
    \centering

    \begin{subfigure}[t]{0.30\textwidth}
        \centering
        \includegraphics[width=\textwidth]{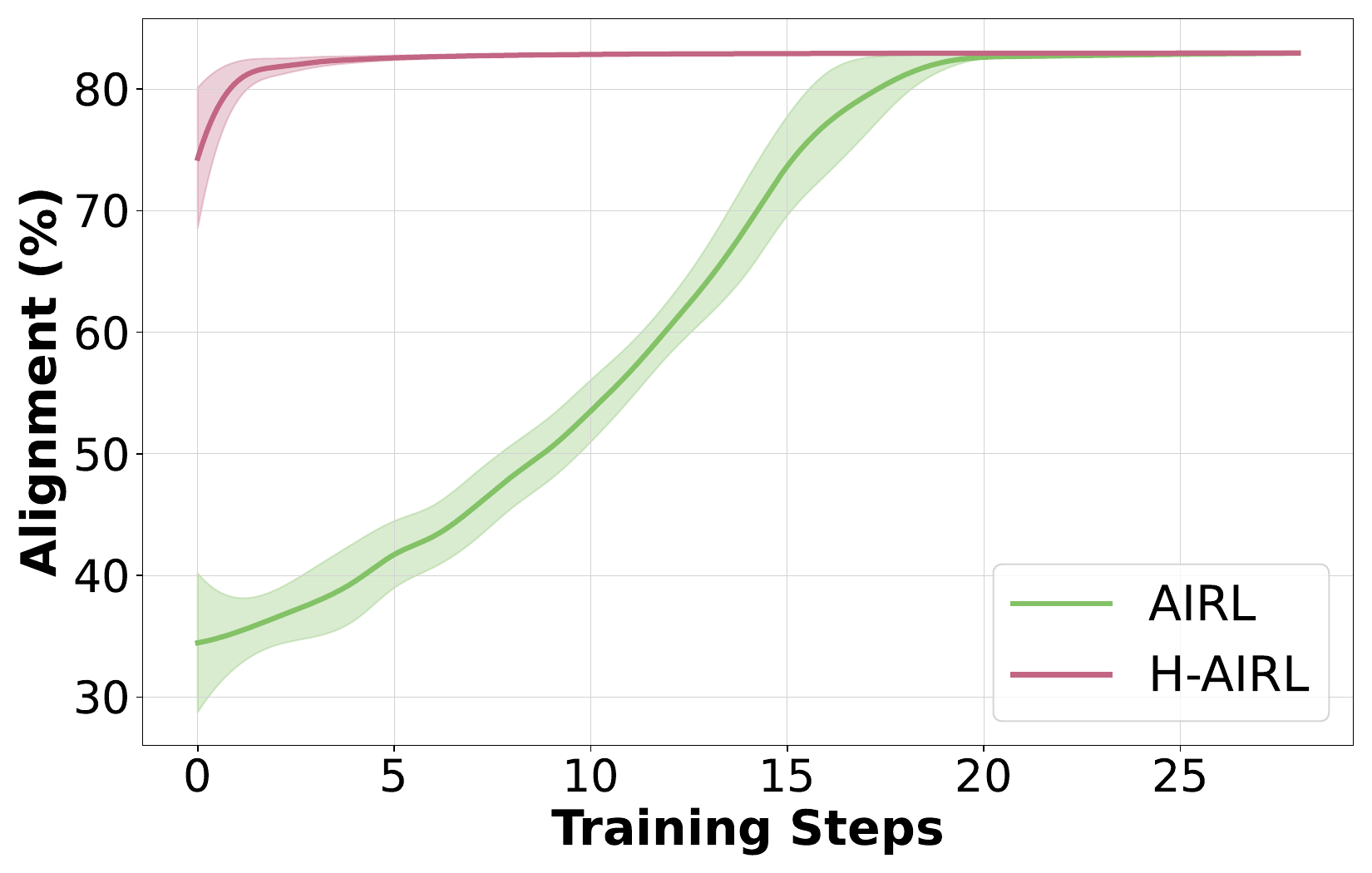}
        \caption{\small Acrobot}
        \label{fig:results:irl:alignments:acrobot}
    \end{subfigure}
    \hspace{1ex}
    \begin{subfigure}[t]{0.30\textwidth}
        \centering
        \includegraphics[width=\textwidth]{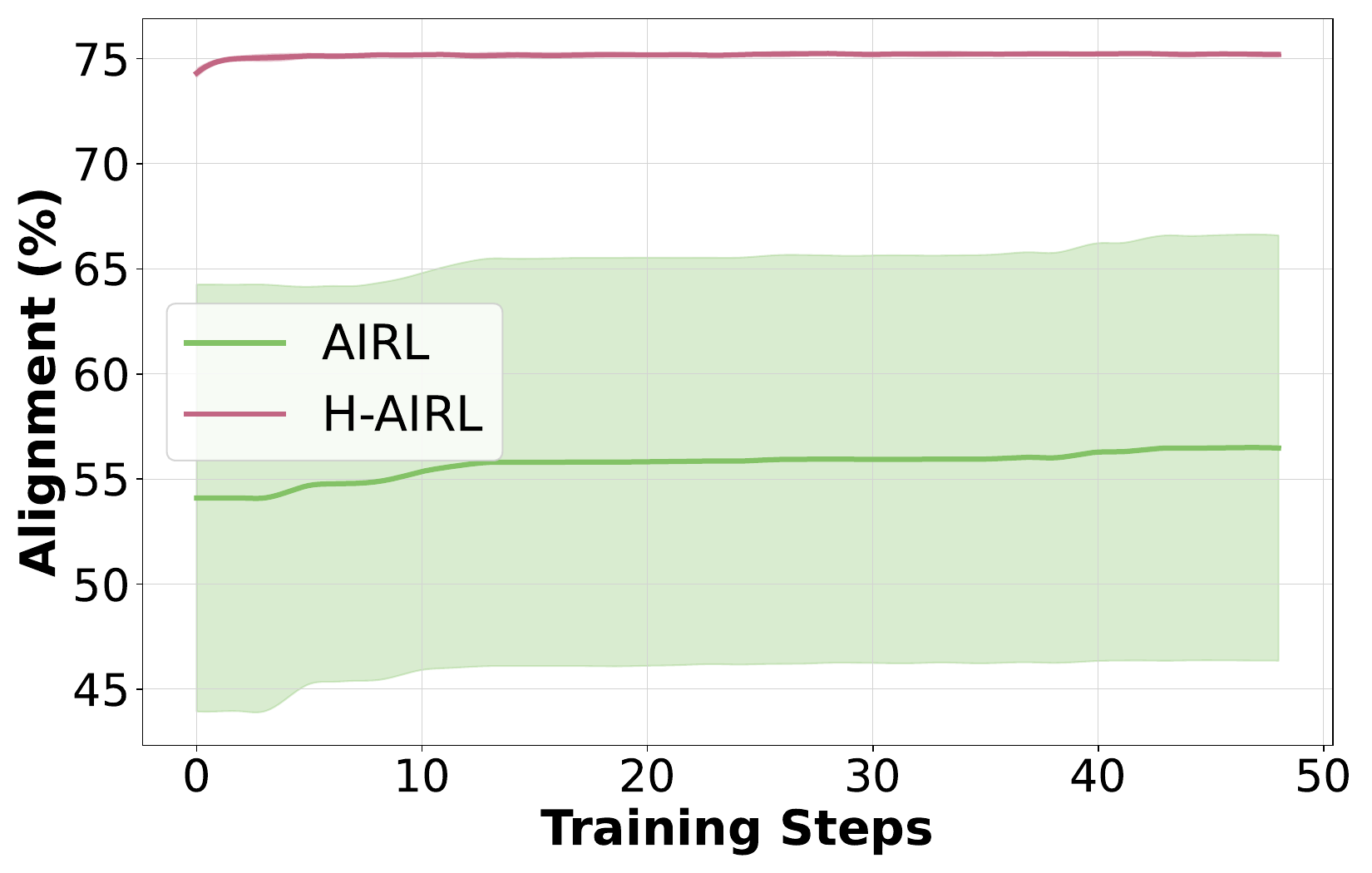}
        \caption{\small LunarLander}
        \label{fig:results:irl:alignments:lunarlander}
    \end{subfigure}

    \vspace{2ex}

    \begin{subfigure}[t]{0.30\textwidth}
        \centering
        \includegraphics[width=\textwidth]{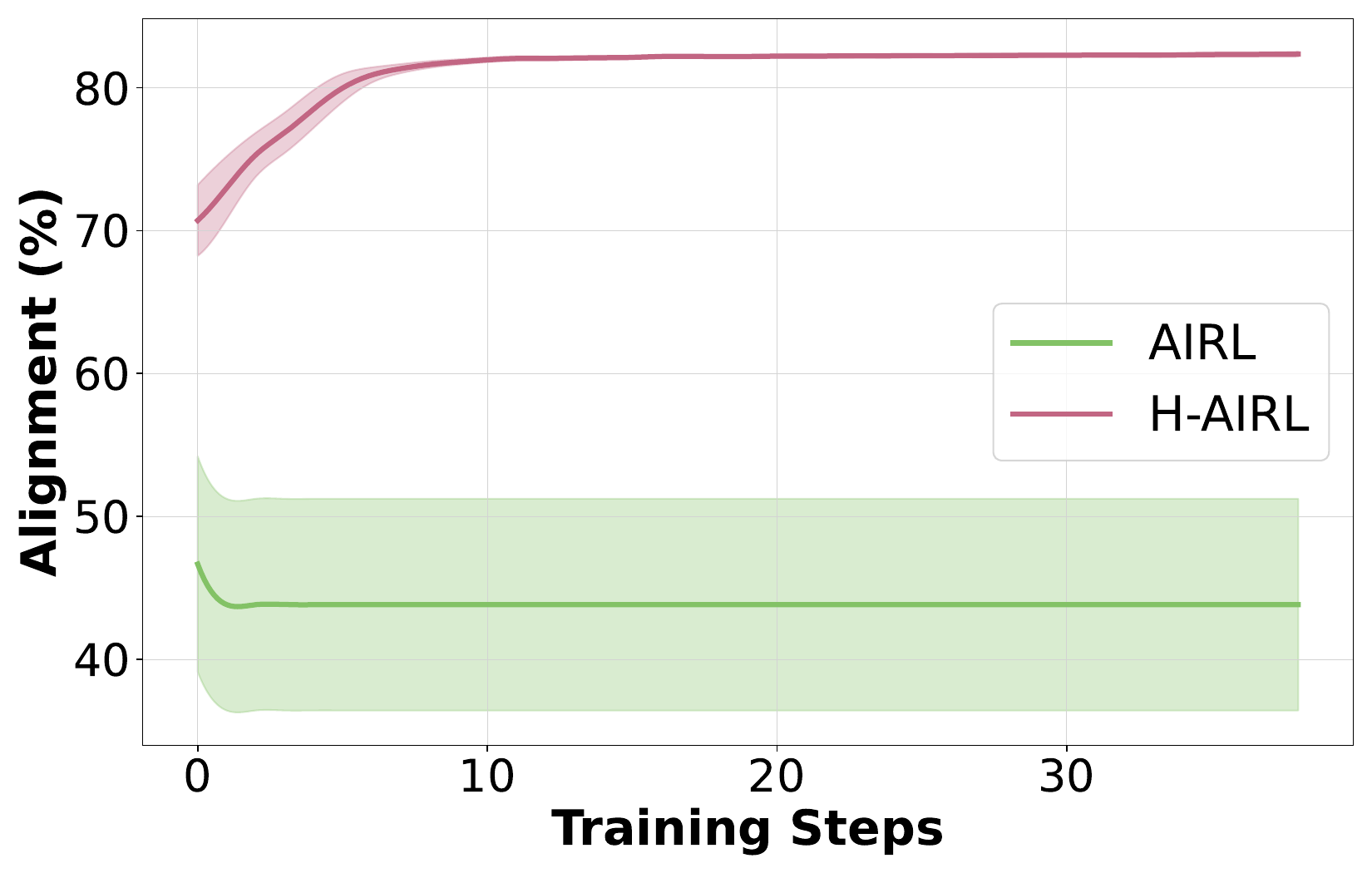}
        \caption{\small MountainCar}
        \label{fig:results:irl:alignments:mountaincar}
    \end{subfigure}
    \hspace{1ex}
    \begin{subfigure}[t]{0.30\textwidth}
        \centering
        \includegraphics[width=\textwidth]{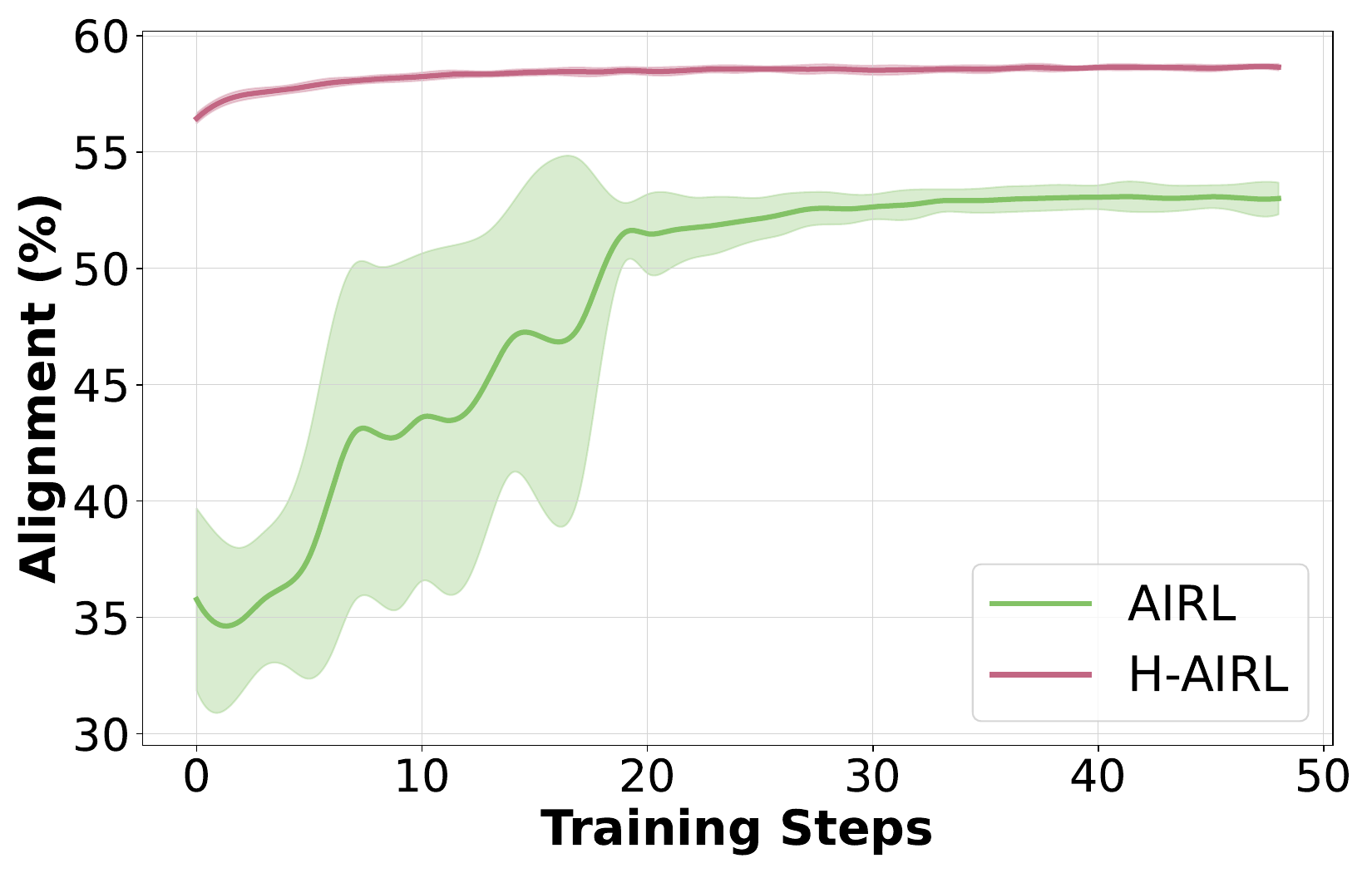}
        \caption{\small HULHE poker}
        \label{fig:results:irl:alignments:poker}
    \end{subfigure}

    \captionsetup{width=0.90\textwidth}
    \caption{\small The policy's state-level action alignment with the expert, for AIRL (green) and H-AIRL (red), across benchmarks with discrete action spaces.}
    \label{fig:results:irl:alignments}
\end{figure}

For tasks with discrete action spaces, including poker, Figure \ref{fig:results:irl:alignments} depicts the policy's state-level action alignment with respect to the expert. 

The H-AIRL policy achieves substantially better action alignment throughout learning as it approaches the expert behavior more quickly and accurately than AIRL, with reduced variance throughout the learning process.

Across both evaluation metrics, reward learning and state-level alignment, the H-AIRL model demonstrates a significantly improved ability to approximate expert-like behavior. It converges more rapidly, exhibits lower variance, and better aligns with expert strategies compared to AIRL.

\subsection{RL Training}\label{sec:results:rl}
We depict the performance of RL agents that utilize the learned reward functions from AIRL and H-AIRL. 

\begin{figure}[H]
    \centering

    \begin{subfigure}{0.32\textwidth}
        \centering
        \includegraphics[width=\linewidth]{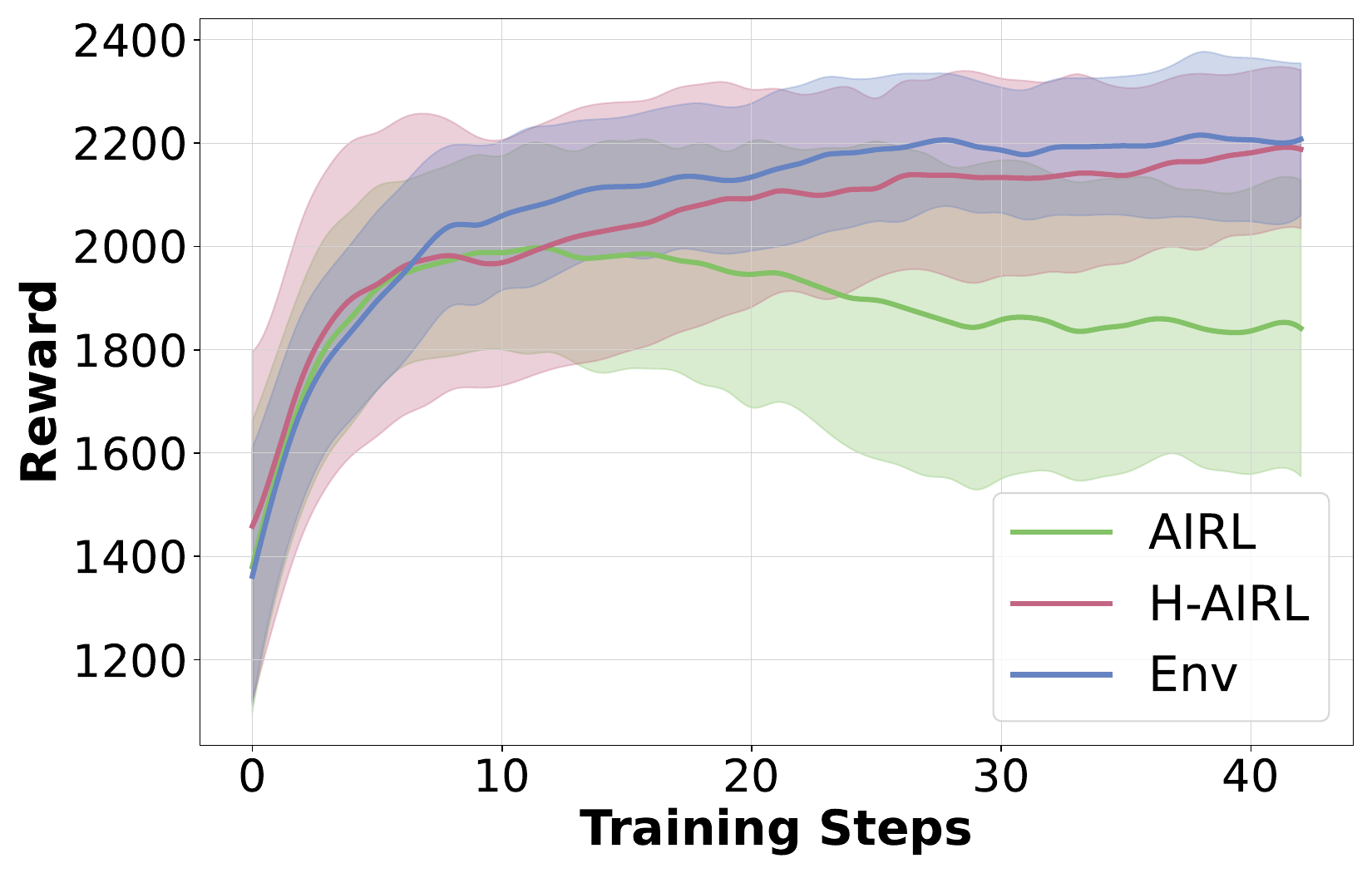}
        \caption{\small HULHE poker}
        \label{fig:results:rl:rewards:poker}
    \end{subfigure}
    \hfill
    \begin{subfigure}{0.32\textwidth}
        \centering
        \includegraphics[width=\linewidth]{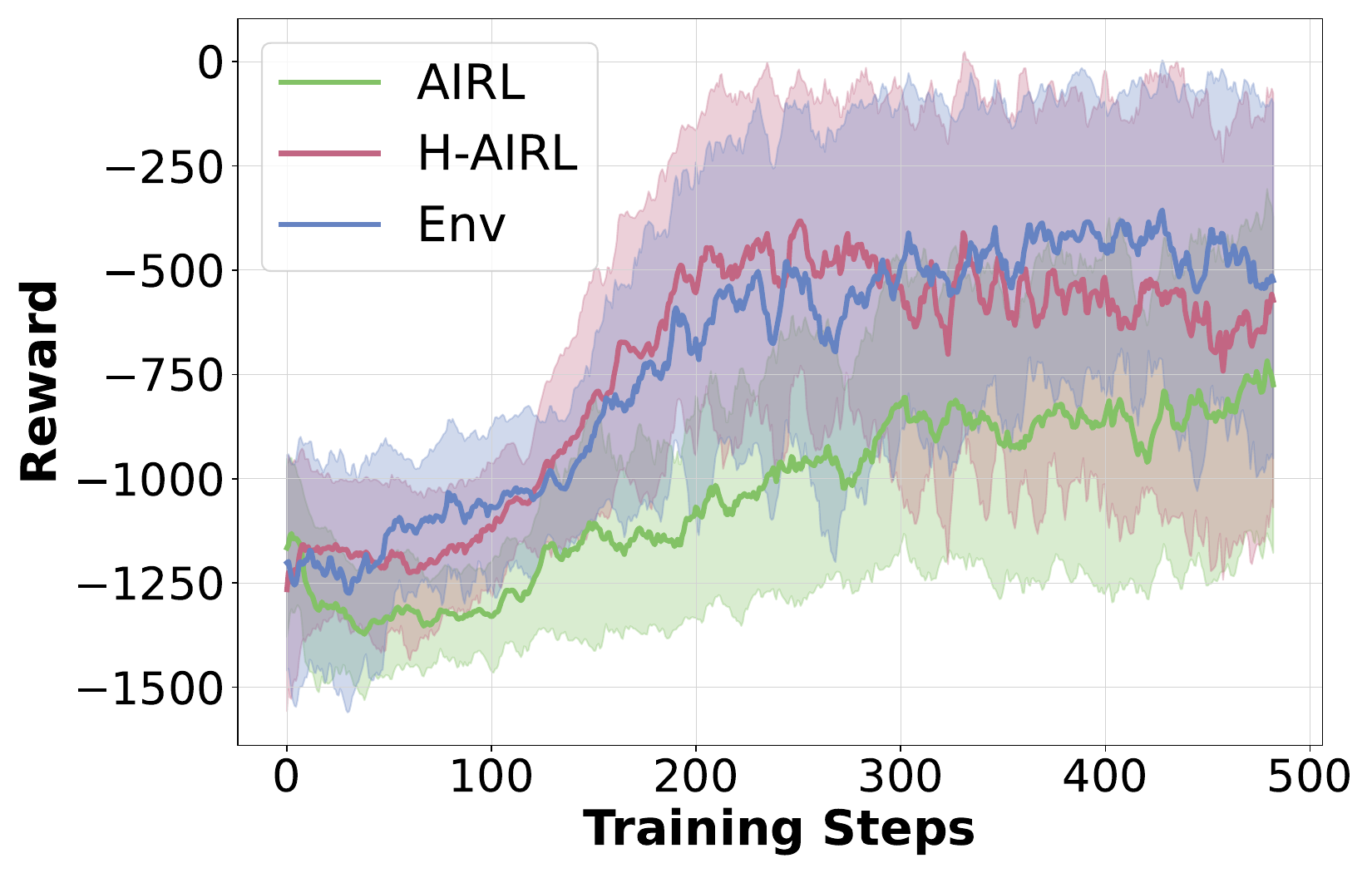}
        \caption{\small Pendulum}
        \label{fig:results:rl:rewards:pendulum}
    \end{subfigure}
    \hfill
    \begin{subfigure}{0.32\textwidth}
        \centering
        \includegraphics[width=\linewidth]{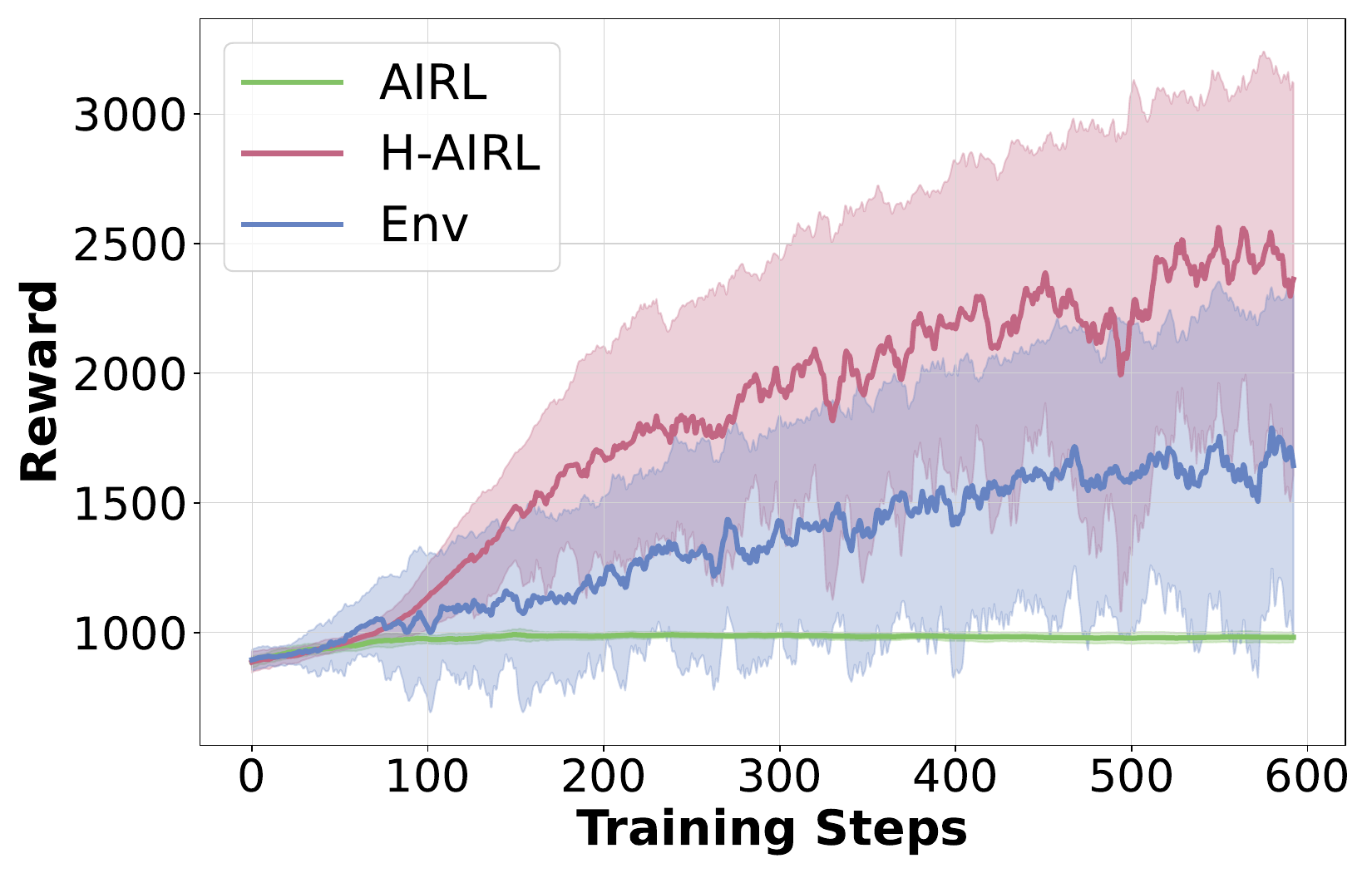}
        \caption{\small Ant}
        \label{fig:results:rl:rewards:ant}
    \end{subfigure}

    \vspace{2ex}

    \begin{subfigure}{0.32\textwidth}
        \centering
        \includegraphics[width=\linewidth]{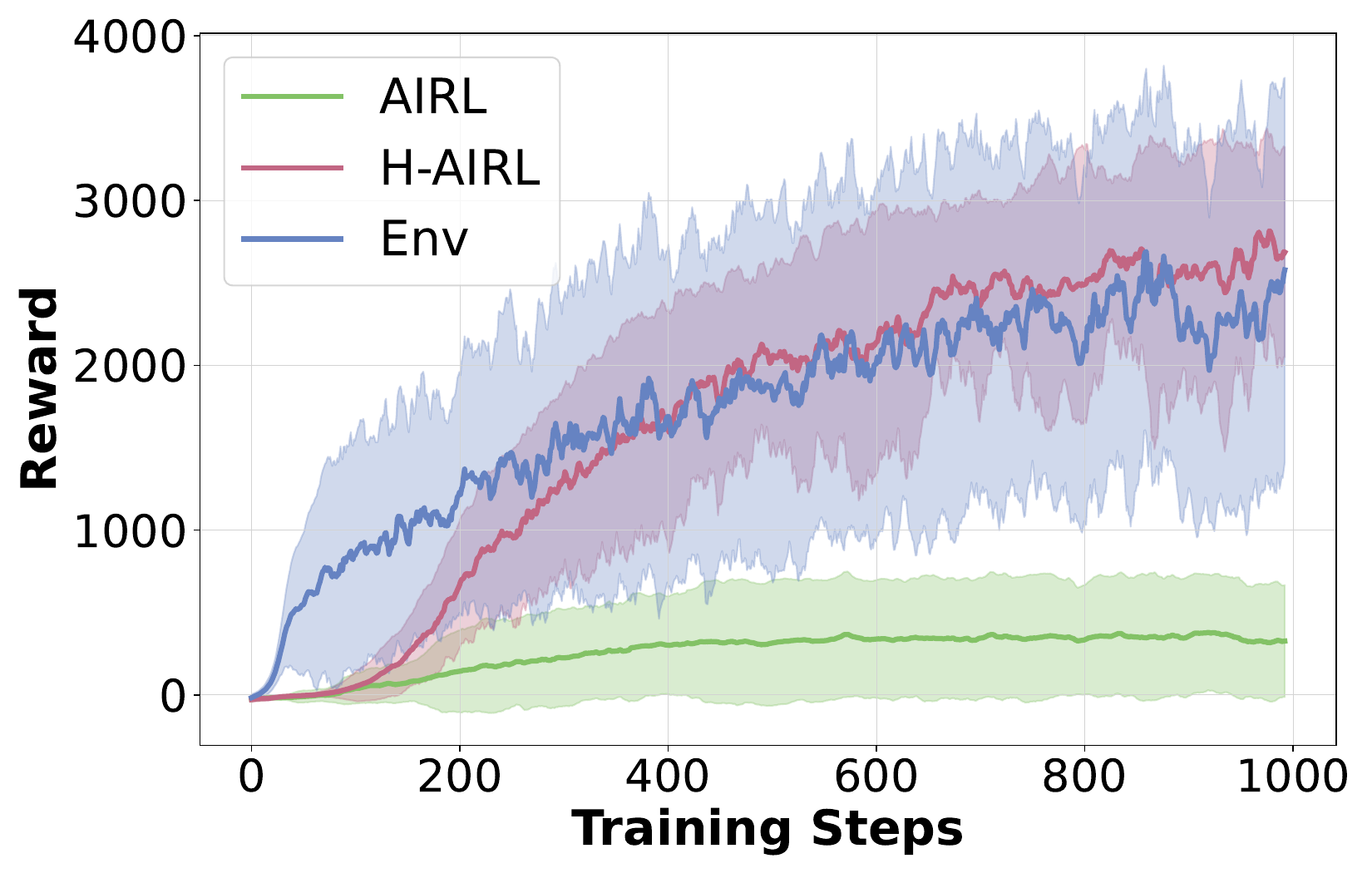}
        \caption{\small HalfCheetah}
        \label{fig:results:rl:rewards:halfcheetah}
    \end{subfigure}
    \hfill
    \begin{subfigure}{0.32\textwidth}
        \centering
        \includegraphics[width=\linewidth]{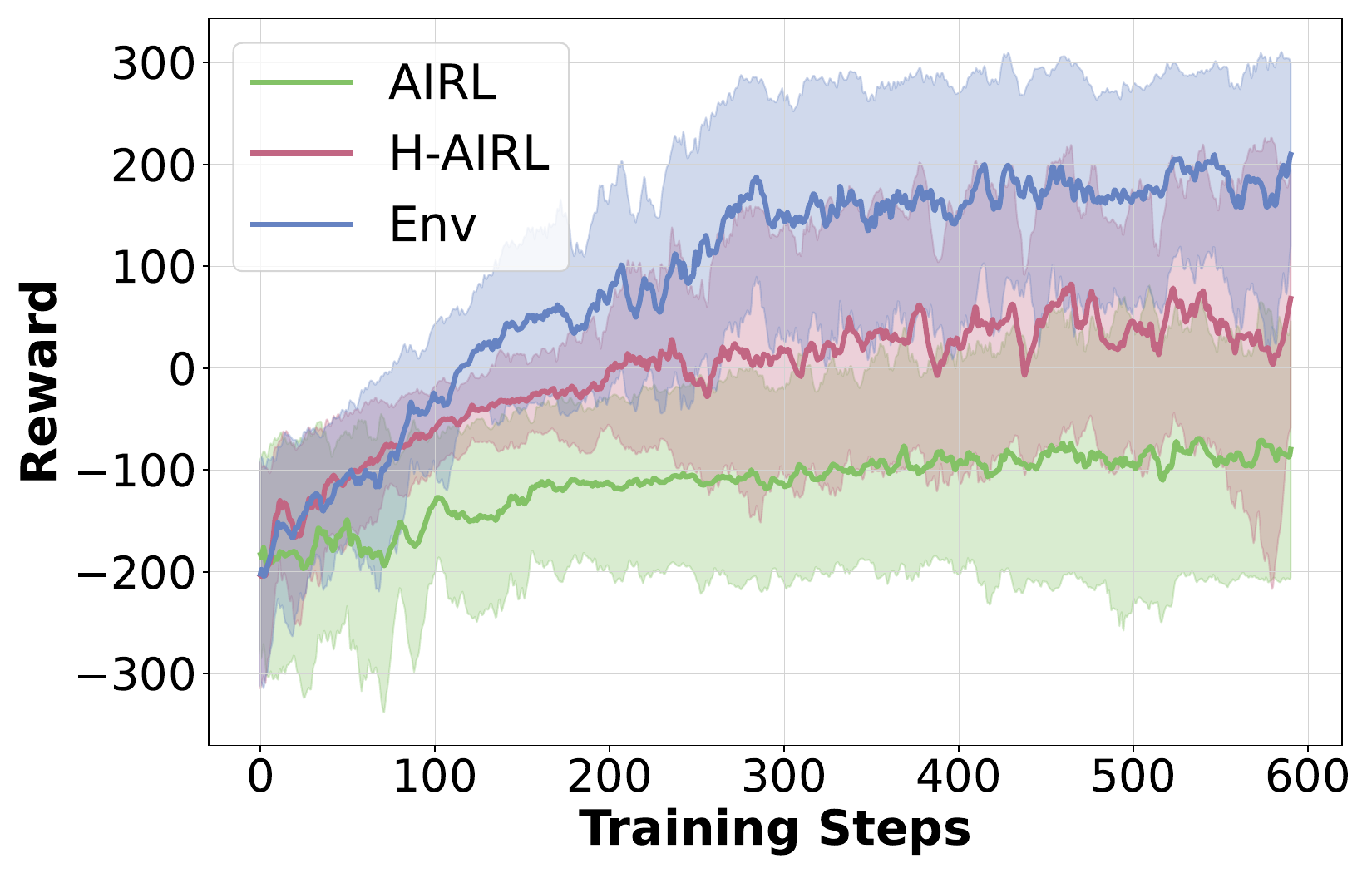}
        \caption{\small LunarLander}
        \label{fig:results:rl:rewards:lunarlander}
    \end{subfigure}
    \hfill
    \begin{subfigure}{0.32\textwidth}
        \centering
        \includegraphics[width=\linewidth]{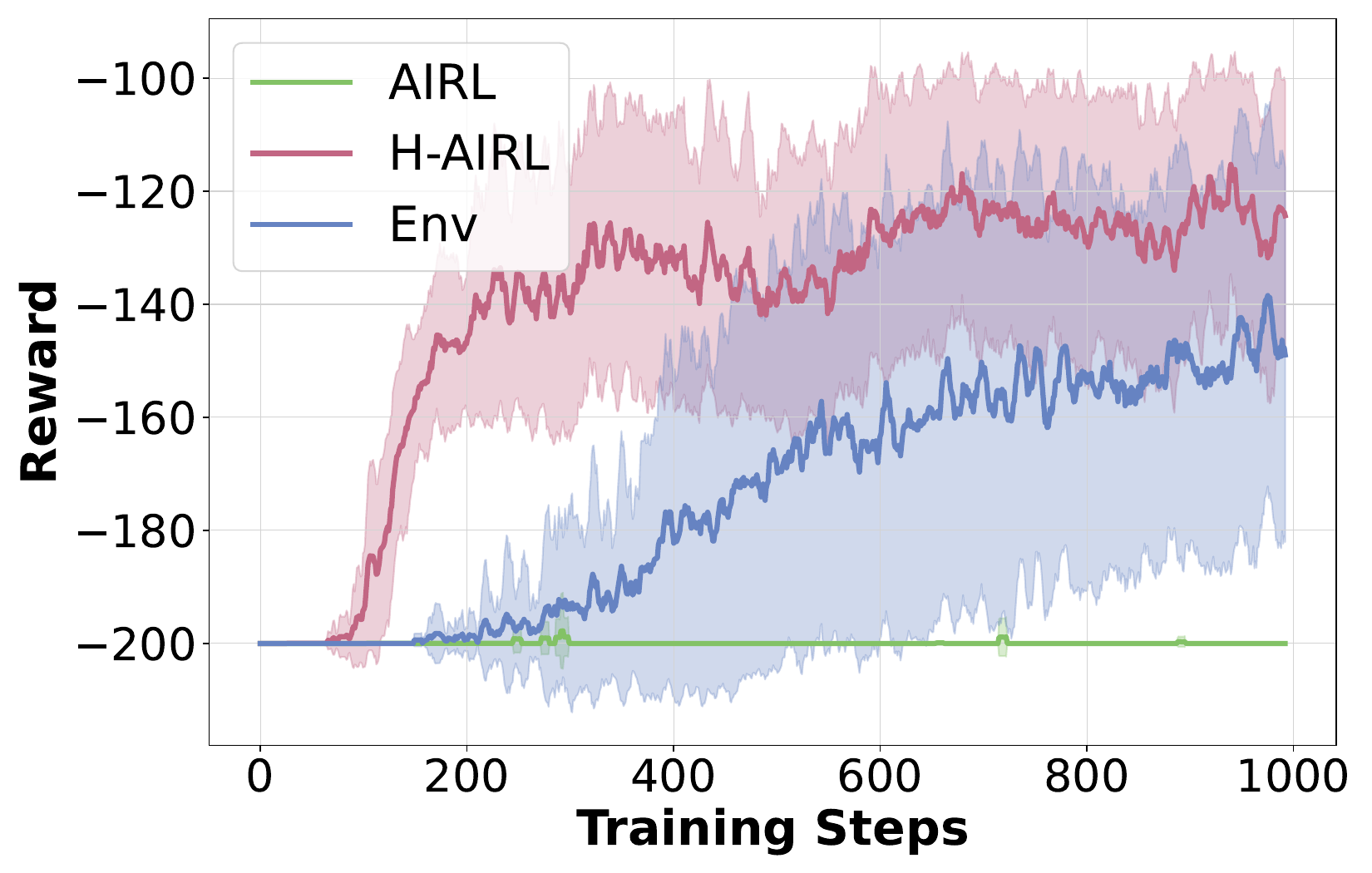}
        \caption{\small MountainCar}
        \label{fig:results:rl:rewards:mountaincar}
    \end{subfigure}

    \captionsetup{width=0.90\textwidth}
    \caption{\small RL training curves of PPO or DQN agents using environment (blue), AIRL-derived (green), and H-AIRL-derived (red) rewards on Gymnasium benchmarks and Heads-Up Limit Hold'em poker.}
    \label{fig:results:rl:rewards}
\end{figure}

Figure \ref{fig:results:rl:rewards} shows the learning curves of an RL agent (PPO or DQN) on the Gymnasium benchmarks, using the learned reward function from the IRL training phase. Across all tasks, we notice a better or otherwise equal ability of H-AIRL to approach expert learning compared to AIRL.

\begin{figure}[H]
    \vspace{2ex}
    \centering
    \begin{subfigure}[t]{0.35\textwidth}
        \centering
        \includegraphics[width=\textwidth]{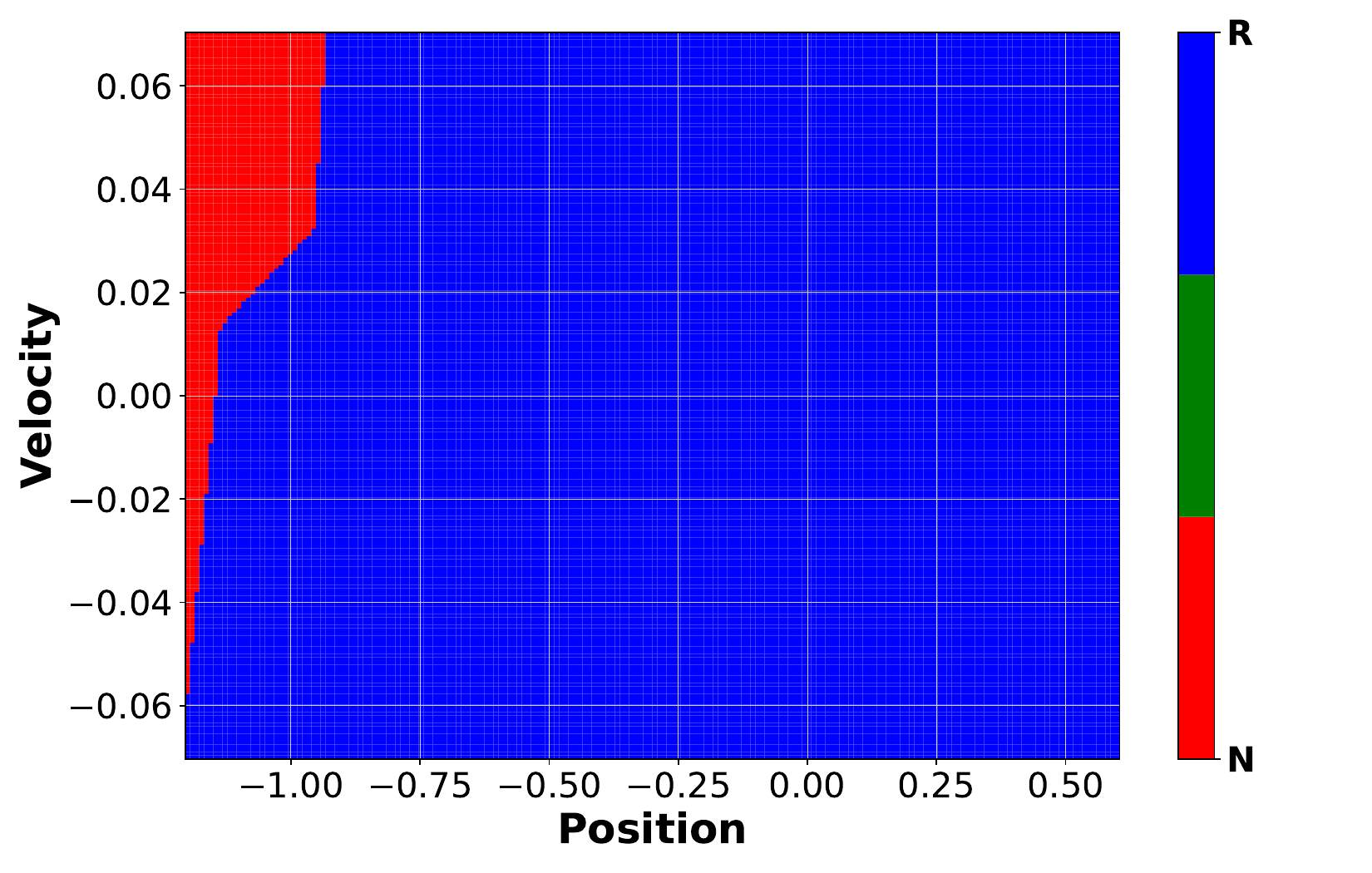}
        \caption{\small AIRL}
        \label{fig:results:rl:reward_function:airl}
    \end{subfigure}
    \hspace{1ex}
    \begin{subfigure}[t]{0.35\textwidth}
        \centering
        \includegraphics[width=\textwidth]{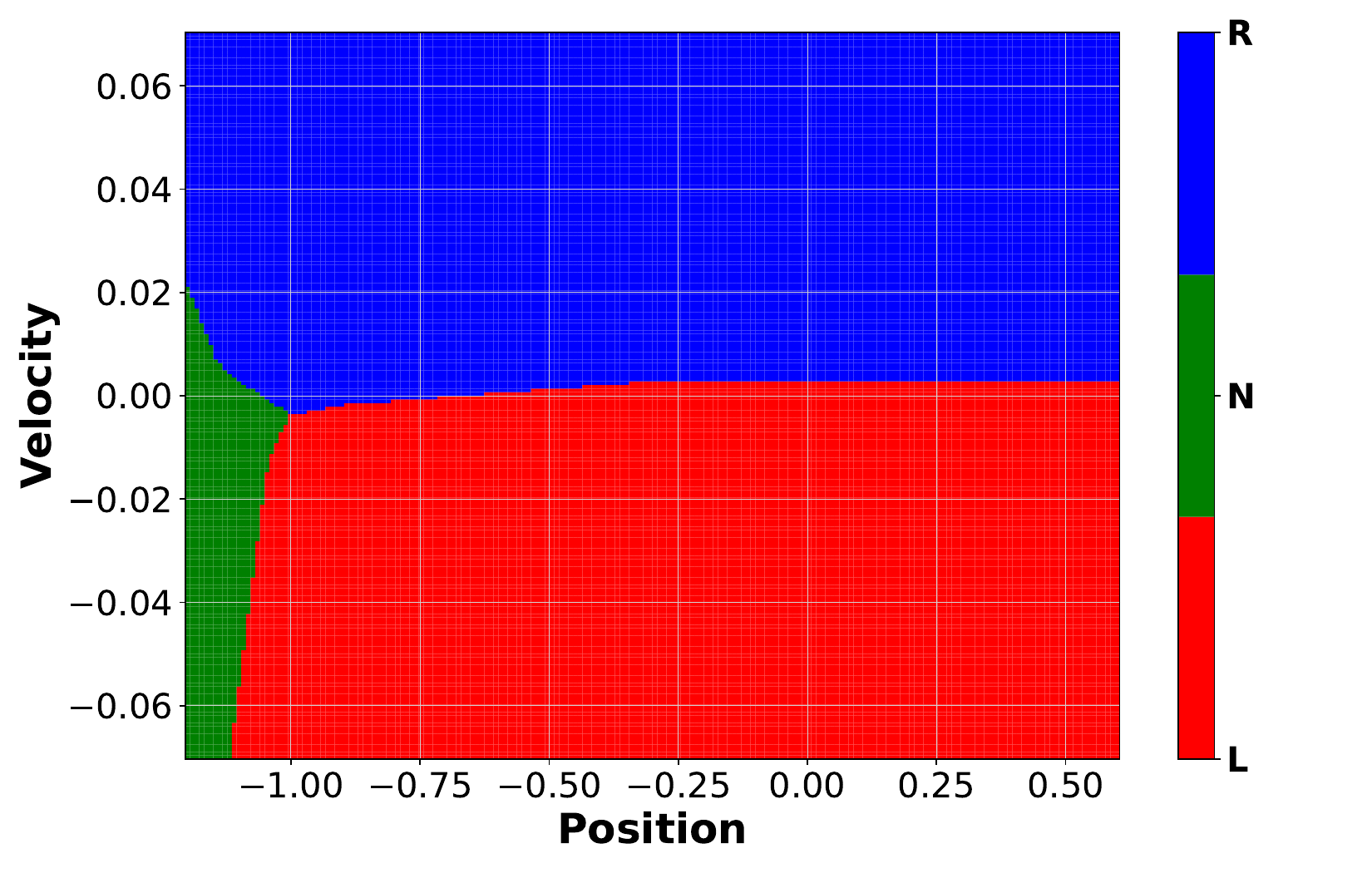}
        \caption{\small H-AIRL}
        \label{fig:results:rl:reward_function:hairl}
    \end{subfigure}

    \captionsetup{width=0.90\textwidth}
    \caption{\small Preferred actions according to the learned reward functions over the MountainCar state space (position vs.\ velocity), for each discrete action: ``thrust right'' (R, blue), ``no thrust'' (N, green), or ``thrust left'' (L, red).}
    \label{fig:results:rl:reward_function}
\end{figure}

This difference in the learned reward functions can be illustrated further by visualizing the reward function output in MountainCar, as shown in Figure \ref{fig:results:rl:reward_function}. In Figure \ref{fig:results:rl:reward_function:airl}, we see that AIRL's learned reward has a strong bias towards the ``thrust right'' action, and there is no area of the state space where ``no thrust'' is the most valuable action in terms of reward. As a result, the agent never learns the back-and-forth switching between actions that is needed to build enough momentum to complete the task successfully. By contrast, in Figure \ref{fig:results:rl:reward_function:hairl}, we see that H-AIRL's reward function is well-balanced, having appropriate areas of the state-action space where each action is most valuable, allowing the RL agent to reproduce the expert's oscillatory strategy and reach the goal.

For poker, Table \ref{tab:results:tournaments} provides the results of 1,000,000 tournaments of AIRL-DQN and H-AIRL-DQN against RLCard's default DQN model to see whether agents trained with a learned reward function can beat agents trained with the traditional (sparse) reward of just the game's payout. To account for potential correlations between seeds, we apply the Bonferroni correction \cite{abdi2007bonferroni}, adjusting the significance threshold to $p < 0.0025$. 

\begin{table}[H]
    \vspace{1ex}
    \centering
        \begin{tabular}{l l l l l}
            \hline
            \noalign{\vspace{0.5ex}}
            $\text{Model}\qquad$  & $\text{Payoff (mbb/h)}\qquad$  & $\text{p-value}\qquad$  \\
            \noalign{\vspace{0.5ex}}
            \hline
            \hline
            \noalign{\vspace{1.5ex}}
            H-AIRL-DQN            & $+96 \pm 14$                   & $<10^{-10}$   \\
            AIRL-DQN              & $-693 \pm 34$                  & $<10^{-10}$   \\
            \noalign{\vspace{1ex}}
            \hline
            \vspace{0pt}
        \end{tabular}

        \captionsetup{width=0.90\textwidth}
        \caption{\small The performance (i.e., the average payoff and standard error in mbb/h) of AIRL-DQN and H-AIRL-DQN against DQN in HULHE poker.}
    \label{tab:results:tournaments}
    \vspace{-2ex}
\end{table}

The tournament results presented in Table \ref{tab:results:tournaments} show a stark contrast between AIRL-DQN and H-AIRL-DQN when competing against RLCard's default DQN agent. AIRL-DQN performs significantly worse than DQN, yielding a negative payoff of $-693 \pm 34$ mbb/h, indicating that it consistently loses to the baseline. In contrast, H-AIRL-DQN achieves a positive payoff of $+96 \pm 14$ mbb/h, demonstrating that it not only outperforms AIRL-DQN, but also performs competitively with the default DQN model. This difference is significant, considering that professional poker players consider 50 mbb/h a sizable margin \cite{moravcik2017deepstack}.

\section{Ablation Study}\label{sec:hyperparams}
Hybrid-AIRL introduces four core hyperparameters:

\begin{itemize}[leftmargin=1.5em, itemsep=0.75em]
    \item \textbf{$\boldsymbol{\alpha}\!\in[0,1]$} is the \textit{policy supervision weight}, which scales the supervised cross-entropy term in the policy loss, as defined in Equation \eqref{eq:h-airl:hybrid_formulation}.  
    \item \textbf{$\boldsymbol{\beta}\!\in[0,1]$} is the \textit{discriminator supervision weight}, which blends the mean-squared error against ground-truth rewards into the discriminator loss, as defined in Equation \eqref{eq:h-airl:disc_h-airl}.  
    \item \textbf{$\boldsymbol{\sigma_{\text{start}}}\!\in[0,1]$} defines the normalized \textit{initial noise standard deviation}, which is the standard deviation of the Gaussian perturbation applied to the first sample in every mini-batch, as described in Section \ref{sec:h-airl:noise}.  
    \item \textbf{$\boldsymbol{\sigma_{\text{end}}}\!\in[0,1]$} defines the normalized \textit{final noise standard deviation}, which is applied to the last sample in the mini-batch; values $\sigma_{\text{end}}>0$ leave a residual noise floor.
\end{itemize}

Collectively, these hyperparameters control the interplay between adversarial learning, supervised learning, and stochastic regularization. Because their effects are interdependent, it is important to understand how each component contributes to the IRL and RL training dynamics. We therefore study these factors individually through a set of controlled ablation experiments.

\subsection{Effect of Policy Supervision \texorpdfstring{($\alpha$)}{alpha}} 
\begin{figure}[H]
    \centering
    \begin{subfigure}[t]{0.30\textwidth}
        \centering
        \includegraphics[width=\textwidth]{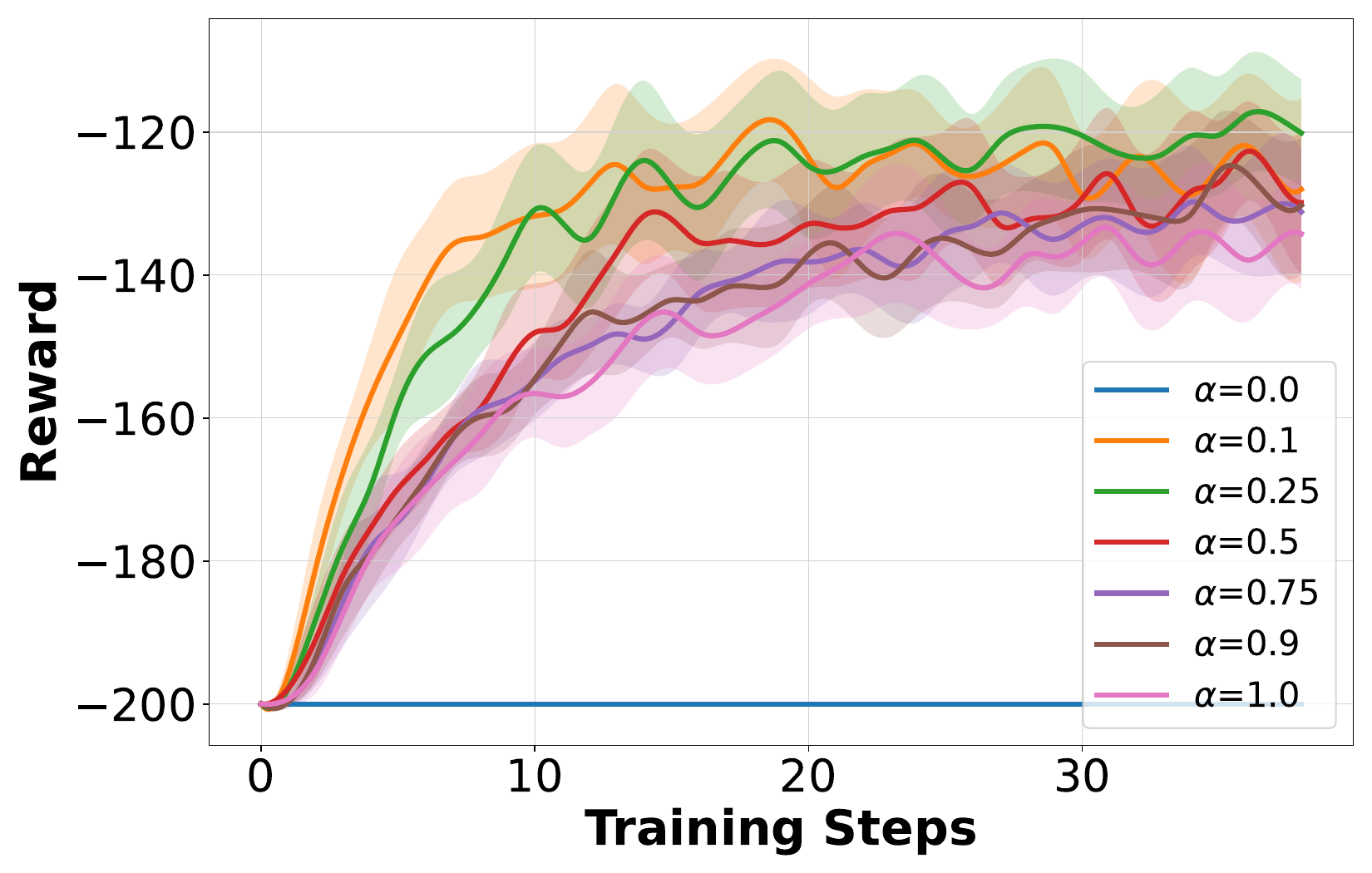}
        \caption{\small Effect of $\alpha$ on IRL}
        \label{fig:hp_alpha}
    \end{subfigure}
    \hspace{1ex}
    \begin{subfigure}[t]{0.30\textwidth}
        \centering
        \includegraphics[width=\textwidth]{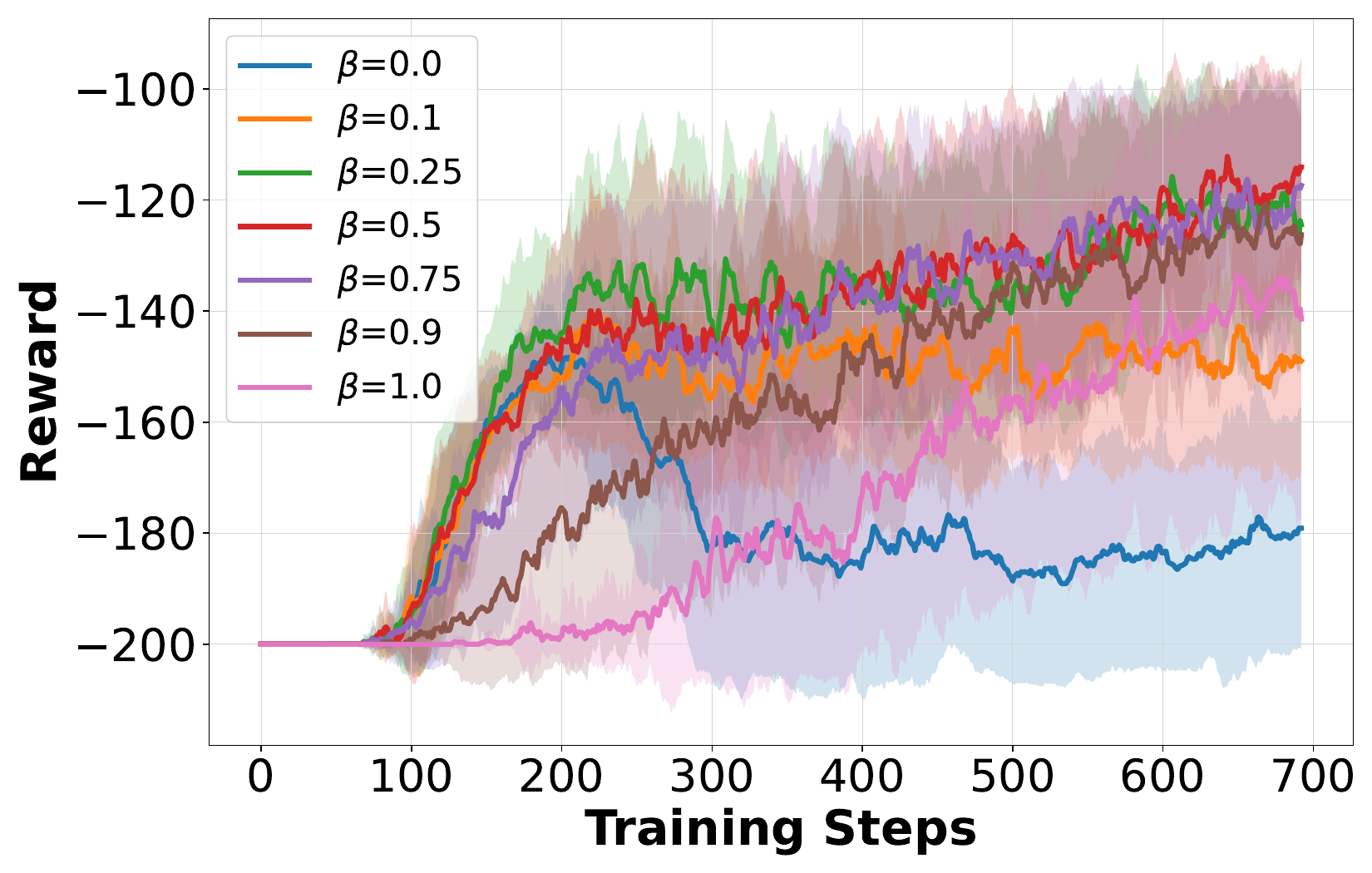}
        \caption{\small Effect of $\beta$ on RL}
        \label{fig:hp_beta}
    \end{subfigure}

    \vspace{2ex}

    \begin{subfigure}[t]{0.30\textwidth}
        \centering
        \includegraphics[width=\textwidth]{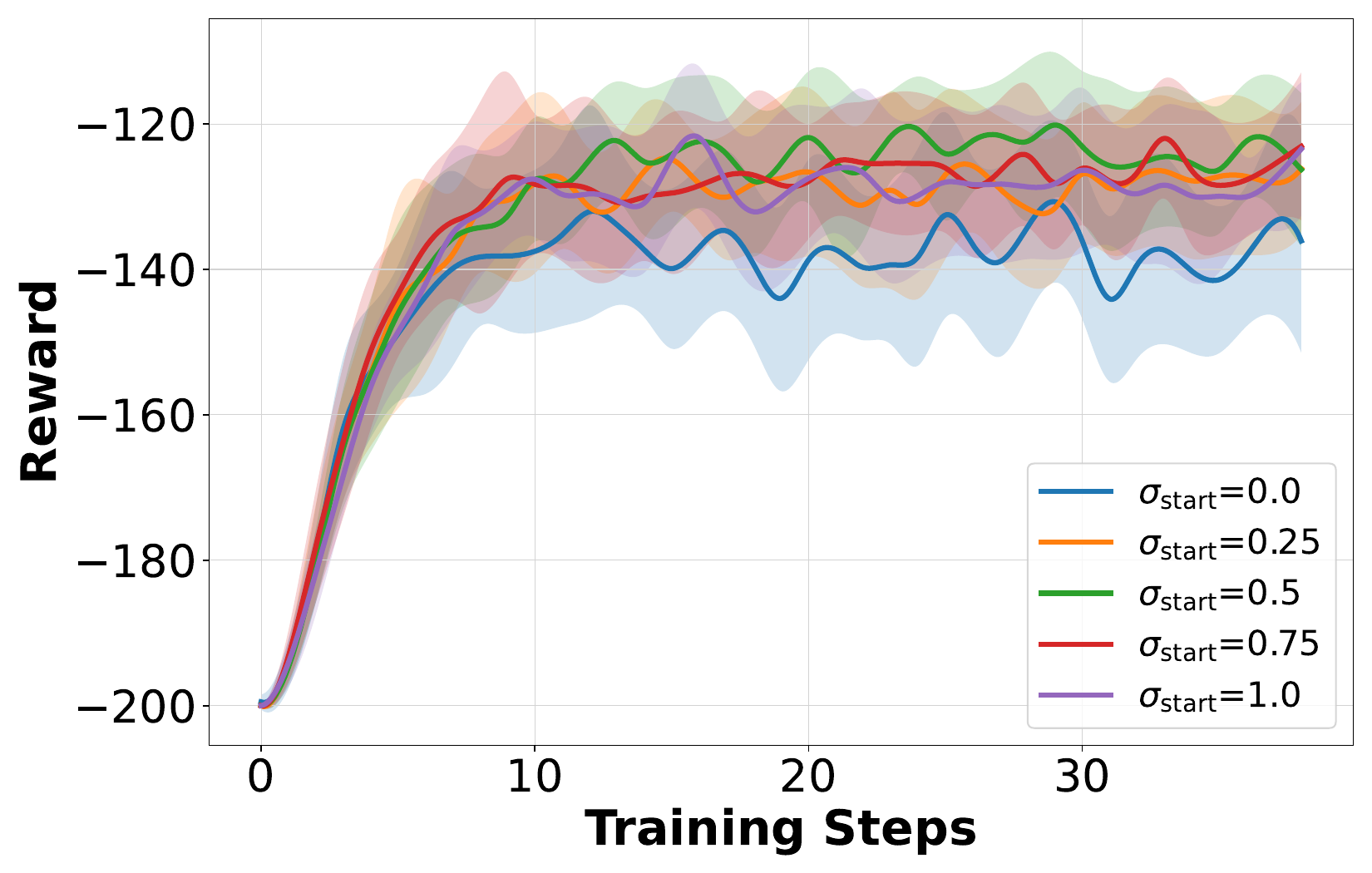}
        \caption{\small Effect of $\sigma_{\text{start}}$ on IRL}
        \label{fig:hp_sigma_start}
    \end{subfigure}
    \hspace{1ex}
    \begin{subfigure}[t]{0.30\textwidth}
        \centering
        \includegraphics[width=\textwidth]{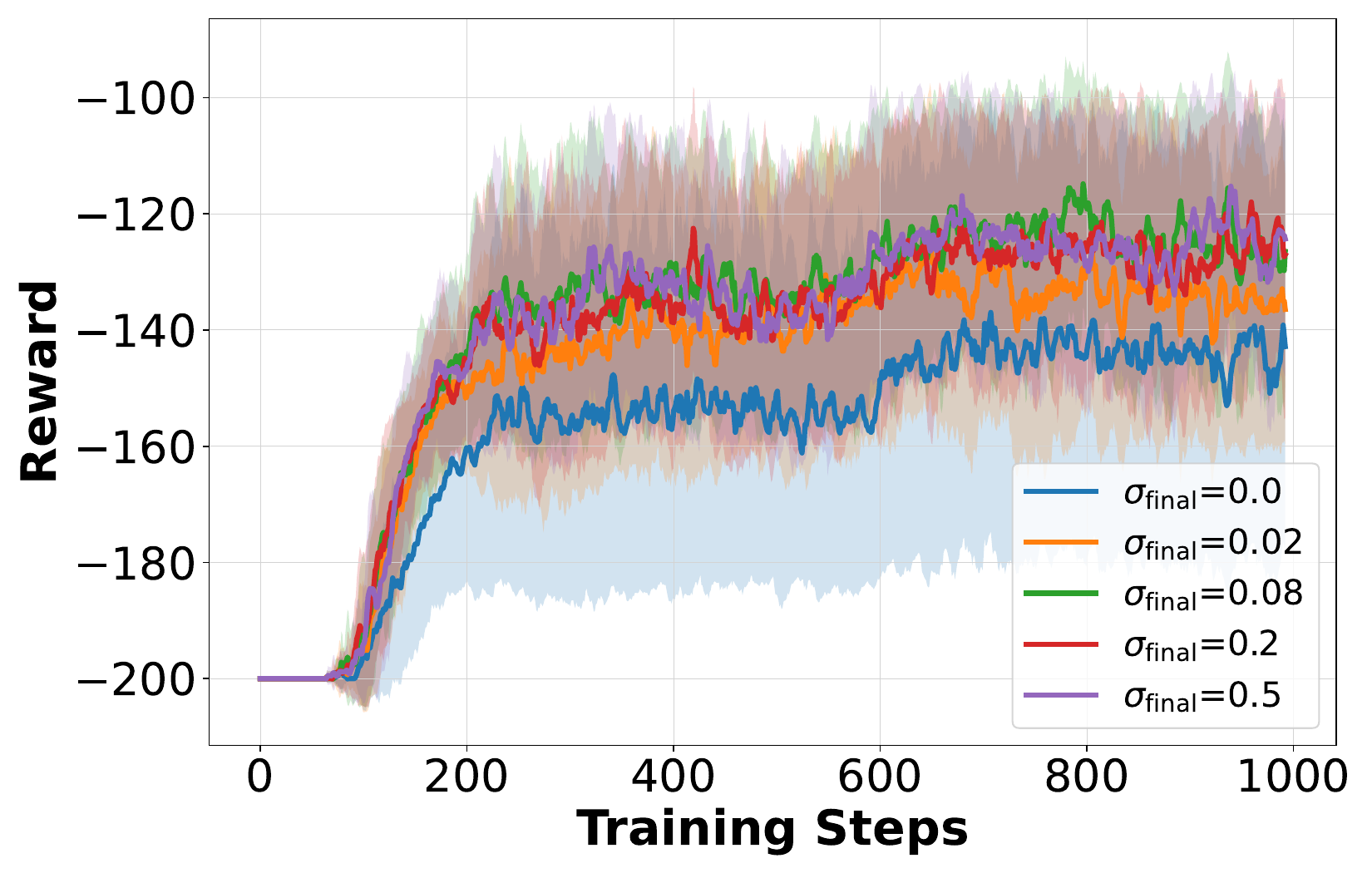}
        \caption{\small Effect of $\sigma_{\text{end}}$ on RL}
        \label{fig:hp_sigma_end}
    \end{subfigure}

    \captionsetup{width=0.90\textwidth}
    \caption{\small One‐factor‐at‐a‐time (OFAT) sweeps on MountainCar for H‐AIRL's core hyperparameters: (a) the policy supervision weight $\alpha$, (b) the discriminator supervision weight $\beta$, (c) the initial noise standard deviation $\sigma_{\text{start}}$, and (d) the final noise standard deviation $\sigma_{\text{end}}$. Each curve shows the mean performance and standard deviation over 10 independent runs.}
    \label{fig:hp_study}
\end{figure}

Figure \ref{fig:hp_alpha} plots the learning curve for MountainCar as $\alpha$ varies during the IRL training phase. Introducing a relatively small amount of policy supervision (e.g., $\alpha \approx 0.1$) provides sufficient guidance for the generator to identify the expert-like action distribution, resulting in the substantial performance gains over AIRL observed in Section \ref{sec:results}. However, full supervision (i.e., $\alpha=1$) hurts performance. With $\alpha=1$ the adversarial game is eliminated, preventing the discriminator from learning a meaningful reward signal.

We note that the value of $\alpha$ that yields the highest policy return may not coincide with the value that results in the best performance when the inferred reward is used to train a separate reinforcement learning agent. This subtle trade-off suggests that designers may prioritize either fast policy convergence or reward fidelity, depending on the application.

\subsection{Effect of Discriminator Supervision \texorpdfstring{($\beta$)}{beta}}
A complementary trend is observed in Figure \ref{fig:hp_beta}. Introducing a moderate level of reward supervision (e.g., $\beta \approx 0.25$) improves the performance of the learned reward function. However, assigning excessively high values, such as $\beta = 1.0$, leads to performance degradation once again. These results reinforce the hypothesis that a hybrid approach of combining both supervised and adversarial learning yields the best performance, whereas using either approach alone is often suboptimal.

While reward supervision can be beneficial, it is not essential for learning a high-performing reward function. For example, our results in the HULHE poker setting were achieved without any reward supervision.

\subsection{Effect of Stochastic Regularization \texorpdfstring{($\sigma_{\text{start}},\sigma_{\text{end}}$)}{sigma-start, sigma-end}}
Noise injection benefits training, especially in more complex environments. Figures \ref{fig:hp_sigma_start} and \ref{fig:hp_sigma_end} show that by starting with substantial noise levels (e.g., $\sigma_{\text{start}}\approx0.9$) and decaying to a smaller nonzero $\sigma_{\text{end}}$, we ensure the discriminator continually sees a diverse spectrum of action qualities, even in late stages of training, thereby avoiding overfitting to homogenized, near‑expert behaviors. Moreover, retaining a noise floor (e.g., $\sigma_{\text{end}}\approx 0.08$) is often better than annealing to zero. A nonzero $\sigma_{\text{end}}$ prevents an ``expert‑vs‑expert'' collapse; by preserving input diversity through perturbations, H-AIRL enables the discriminator to learn robust distinctions rather than fitting to noise.

\section{Discussion}
We experimentally show that H-AIRL consistently outperforms AIRL, achieving improved policy learning and reward function inference across all evaluation metrics. While AIRL eventually learns an effective policy in Gymnasium benchmarks, H-AIRL exhibits faster convergence, greater training stability, better action alignment, and action distributions that better match those of expert players. The advantages of a hybrid IRL framework are further underscored in the context of HULHE poker, where H-AIRL is able to infer a more informative reward function that more effectively guides the RL agent towards expert behavior. This is evidenced by the improved learning curve in the RL training phase and the higher payoffs observed in our tournament evaluations. Overall, these findings suggest that while state-action AIRL serves as a powerful baseline for learning from expert demonstrations, the modifications introduced in H-AIRL provide a robust enhancement that is effective at scaling the IRL approach to even more complex settings. The improved performance of H-AIRL in both Gymnasium benchmarks and HULHE poker highlights its potential as a promising approach for inverse reinforcement learning in real-world domains characterized by highly complex dynamics and sparse rewards.

Our aim in this work is to isolate and evaluate the contribution of H-AIRL's hybrid loss framework relative to the foundational AIRL baseline. A broader empirical evaluation against recent IRL methods is left to future work in order to better situate H-AIRL within the current landscape \cite{zhou2024rethinking, yoon2024maximum}. While our results are encouraging, the study has several limitations. First, our poker data lacks folding actions, a common limitation in real-world datasets where folded cards are never revealed. Second, H-AIRL does not recover disentangled rewards that lead to theoretical guarantees for transfer, and does not explicitly address partial observability. These points suggest several directions for future work, such as extending the hybrid framework formulation to produce disentangled rewards, and studying recurrent or belief-state extensions of H-AIRL for partially observable domains.

\section*{Use of Generative AI}
The use of generative AI was limited to tasks such as language refinement, without replacing critical analysis, original research, or authorship contributions.

\section*{Acknowledgements}
LW and PJKL gratefully acknowledge support from the Research Foundation Flanders (FWO), via ACCELERATE project G059423N. PJKL gratefully acknowledges support from the Research council of the Vrije Universiteit Brussel (OZR-VUB via grant number OZR3863BOF). This research acknowledges funding from the Flemish Government through the AI Research Program. We made use of computational resources and services provided by the Flemish Supercomputer Centre (VSC), funded by the FWO and the Flemish Government.

\bibliographystyle{unsrt}
\bibliography{bibliography}
\end{document}